\documentclass[lettersize,journal]{IEEEtran}
\usepackage{amsmath,amsfonts}
\usepackage[colorlinks,
            linkcolor=red,
            anchorcolor=blue,
            citecolor=green
            ]{hyperref}
\usepackage{hyperref}
\usepackage{algorithm}
\usepackage{array}
\usepackage[caption=false,font=normalsize,labelfont=sf,textfont=sf]{subfig}
\usepackage{textcomp}
\usepackage{stfloats}
\usepackage{url}
\usepackage{verbatim}
\usepackage{graphicx}
\usepackage{cite}
\usepackage{algpseudocode}  
\usepackage[percent]{overpic}
\usepackage{caption}
\usepackage{enumitem}
\usepackage{tikz}
\usepackage{adjustbox}
\usepackage{cleveref}
\usepackage{multirow}
\begin{document}
\title{Simultaneous Image-to-Zero and Zero-to-Noise: Diffusion Models with Analytical Image Attenuation}

\author{Yuhang Huang, Zheng Qin, Xinwang Liu,~\IEEEmembership{Senior Member,~IEEE}, Kai Xu$^{*}$\thanks{$*$ Corresponding author.},~\IEEEmembership{Senior Member,~IEEE}
% \thanks{Y. Huang, X. Liu and K. Xu are with School of Computer, National University of Defense Technology, Changsha, 410073, China (E-mail: {huangai, xinwangliu}\@nudt.edu.cn, kevin.kai.xu@gmail.com). Z. Qin is with Defense Innovation Institute, Academy of Military Sciences, Beijing, 100071, China.
}

\markboth{Journal of \LaTeX\ Class Files,~Vol.~14, No.~8, August~2021}%
{Shell \MakeLowercase{\textit{et al.}}: A Sample Article Using IEEEtran.cls for IEEE Journals}


% \maketitle
% \begin{figure*}[h]
%     \centering
%     % \includegraphics[width=1.\linewidth]{figures/teaser-cvpr.pdf}%\vspace{-10pt}
%     \includegraphics[width=1.\linewidth]{figures/figure1.pdf}%\vspace{-10pt}
%     \caption{High-quality images generated by the proposed DDMs under few-step settings. (a) 10-step unconditioned generation on the CIFAR-10 and CelebA-HQ-256 datasets. (b) 5-step conditioned tasks  %Generated samples on both unconditioned and conditioned tasks. \textbf{(a)}: DDMs can generate high-quality images on CIFAR-10 and CelebA-HQ-256 with only 10 steps. \textbf{(b)}: For conditioned tasks, DDMs only use 5 steps for generating high-quality results 
%     including saliency detection, image inpainting and super-resolution.}%\vspace{-15pt}
%     \label{fig:top}
% \end{figure*}

\twocolumn[{
\renewcommand\twocolumn[1][]{#1}
\maketitle
\begin{center}
    \captionsetup{type=figure}
    \includegraphics[width=\textwidth]{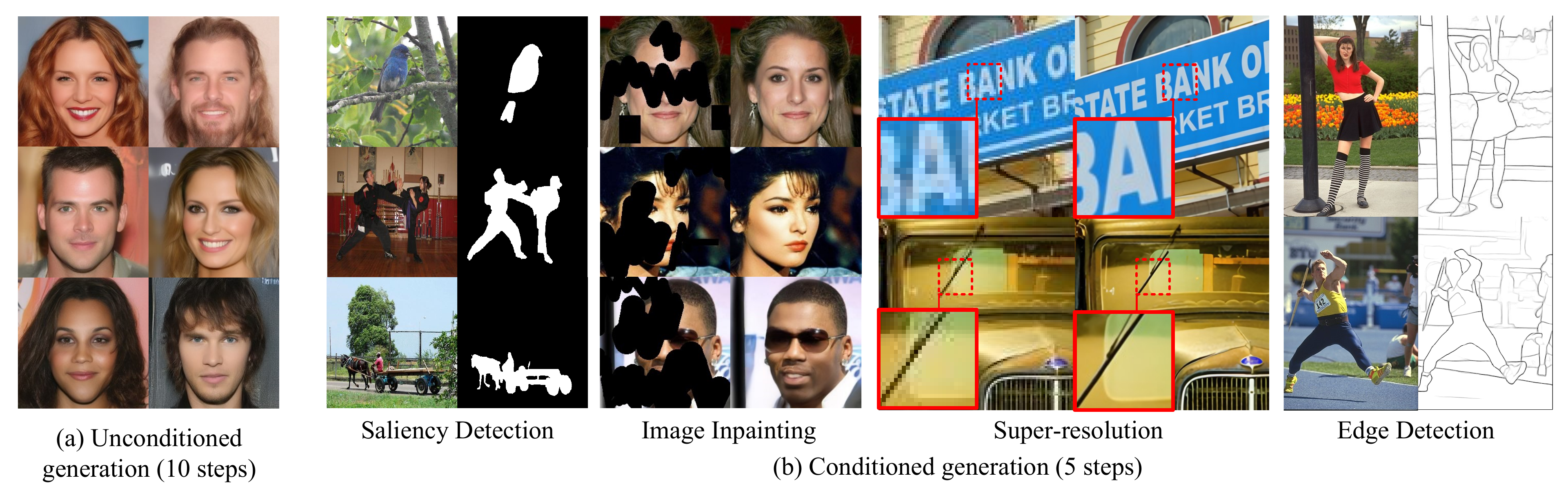}
    \captionof{figure}{High-quality images generated by the proposed ADMs under few-step settings. (a) 10-step unconditioned generation on the CelebA-HQ-256 dataset. (b) 5-step conditioned tasks including saliency detection, image inpainting, super-resolution and edge detection.
    %Generated samples on both unconditioned and conditioned tasks. \textbf{(a)}: DDMs can generate high-quality images on CIFAR-10 and CelebA-HQ-256 with only 10 steps. \textbf{(b)}: For conditioned tasks, DDMs only use 5 steps for generating high-quality results.
    }
\end{center}
}]
{
\renewcommand{\thefootnote}{}
\footnotetext[1]{Corresponding author: Kai Xu (E-mail: kevin.kai.xu@gmail.com).}
\footnotetext[2]{Y. Huang, X. Liu and K. Xu are with School of Computer, National University of Defense Technology, Changsha, 410073, China. Z. Qin is with Defense Innovation Institute, Academy of Military Sciences, Beijing, 100071, China.}
%  (E-mail: {huangai, xinwangliu}\@nudt.edu.cn, kevin.kai.xu@gmail.com)
}
% \saythanks
\begin{abstract}
% Recent diffusion models have shown remarkable performances in image generation, yet they meet challenges in the trade-off between generative quality and speed. 
Recent studies have demonstrated that the forward diffusion process is crucial for the effectiveness of diffusion models in terms of generative quality and sampling efficiency.
% Recent diffusion models have shown remarkable performances in image generation, yet they suffer from the time-consuming sampling process due to thousands of function evaluations. 
% We propose incorporating an analytical image attenuation process into the forward diffusion process for high-quality (un)conditioned image generation in less than 10 rather than vanilla thousands of denoising steps. 
We propose incorporating an analytical image attenuation process into the forward diffusion process for high-quality (un)conditioned image generation with significantly fewer denoising steps compared to the vanilla diffusion model requiring thousands of steps. 
In a nutshell, our method represents the forward image-to-noise mapping as simultaneous \textit{image-to-zero} mapping and \textit{zero-to-noise} mapping. Under this framework, we mathematically derive 1) the training objectives and 
2) for the reverse time the sampling formula based on an analytical attenuation function which models image to zero mapping.  % explicit transition probability · is derived for \textbf{image-to-zero} mapping, we derive the sampling 
The former enables our method to learn noise and image components simultaneously which simplifies learning. Importantly, because of the latter's analyticity in the \textit{zero-to-image} sampling function, we can avoid the ordinary differential equation-based accelerators and instead naturally perform sampling with an arbitrary step size. 
We have conducted extensive experiments on unconditioned image generation, \textit{e.g.}, CIFAR-10 and CelebA-HQ-256, and image-conditioned downstream tasks such as super-resolution, saliency detection, edge detection, and image inpainting.
The proposed diffusion models achieve competitive generative quality with much fewer denoising steps compared to the state of the art, thus greatly accelerating the generation speed. In particular, to generate images of comparable quality, our models require only one-twentieth of the denoising steps compared to the baseline denoising diffusion probabilistic models. Moreover, we achieve state-of-the-art performances on the image-conditioned tasks using only no more than 10 steps.
% Under the few function evaluation setups, the proposed diffusion models experimentally yield very competitive performance compared with the state of the art in 1) unconditioned image generation, \textit{e.g.}, CIFAR-10 and CelebA-HQ-256 and 2) image-conditioned downstream tasks such as super-resolution, saliency detection, edge detection, and image inpainting. 
Code is available at: \href{https://github.com/GuHuangAI/ADM-Public}{https://github.com/GuHuangAI/ADM-Public}.
\end{abstract}


\vspace{-10pt}
\section{Introduction}

\IEEEPARstart{D}{iffusion} probabilistic models (DPMs) \cite{(2)ddpm,(3)pmlr-v139-nichol21a,(5)ho2022classifier, (8)dhariwal2021diffusion} have made impressive achievements in content generation \cite{(4)karras2022elucidating, (6)DBLP:conf/iclr/ChenZZWNC21, (7)DBLP:conf/interspeech/ChenZZW0DC21, (59)10081412, (60)10416192}. DPMs view the image-to-noise process as a parameterized Markov chain that gradually adds noise to the original data until the signal is completely corrupted and uses a neural network to model the reversed denoising process for new sample generation.

%Recent diffusion probabilistic models (DPMs) \cite{(2)ddpm,(3)pmlr-v139-nichol21a,(5)ho2022classifier, (8)dhariwal2021diffusion} have shown state-of-the-art performances in generated content fields such as image and speech synthesis \cite{(4)karras2022elucidating, (6)DBLP:conf/iclr/ChenZZWNC21, (7)DBLP:conf/interspeech/ChenZZW0DC21}, image-to-image translation \cite{(13)sasaki2021unit}, and 3D object generation \cite{(14)luo2021diffusion, (15)poole2022dreamfusion}. Different from variational auto-encoders \cite{(16)DBLP:journals/corr/KingmaW13} or adversarial models \cite{(17)goodfellow2020generative}, DPMs view the image-to-noise process as a parameterized Markov chain that gradually adds noise to the original data until the signal is completely corrupted, and utilize a neural network to model the \emph{reversed denoising process} for new samples generation.

\cite{(9)sde} prove that the forward diffusion process of DPMs is equivalent to the score-based generative models (SGMs) \cite{(10)sohl2015deep, (11)song2021maximum, (12)kingma2021variational}. Both can be represented by a stochastic differential equation (SDE). Essentially, the training of DPMs is to learn the score function, gradient of the log probability density, of the perturbed data. \cite{(9)sde} show that the learned score function is uniquely dependent on the forward diffusion process. Therefore, the forward diffusion process is critical for the effectiveness of DPMs, for example, in terms of generative quality and sampling efficiency.

\begin{figure*}
  \centering
  %\fbox{\rule[-.5cm]{0cm}{4cm} \rule[-.5cm]{4cm}{0cm}}
  %\includegraphics[width=1\linewidth]{figures/framework.pdf}
  \begin{overpic}
    [width=1\textwidth]{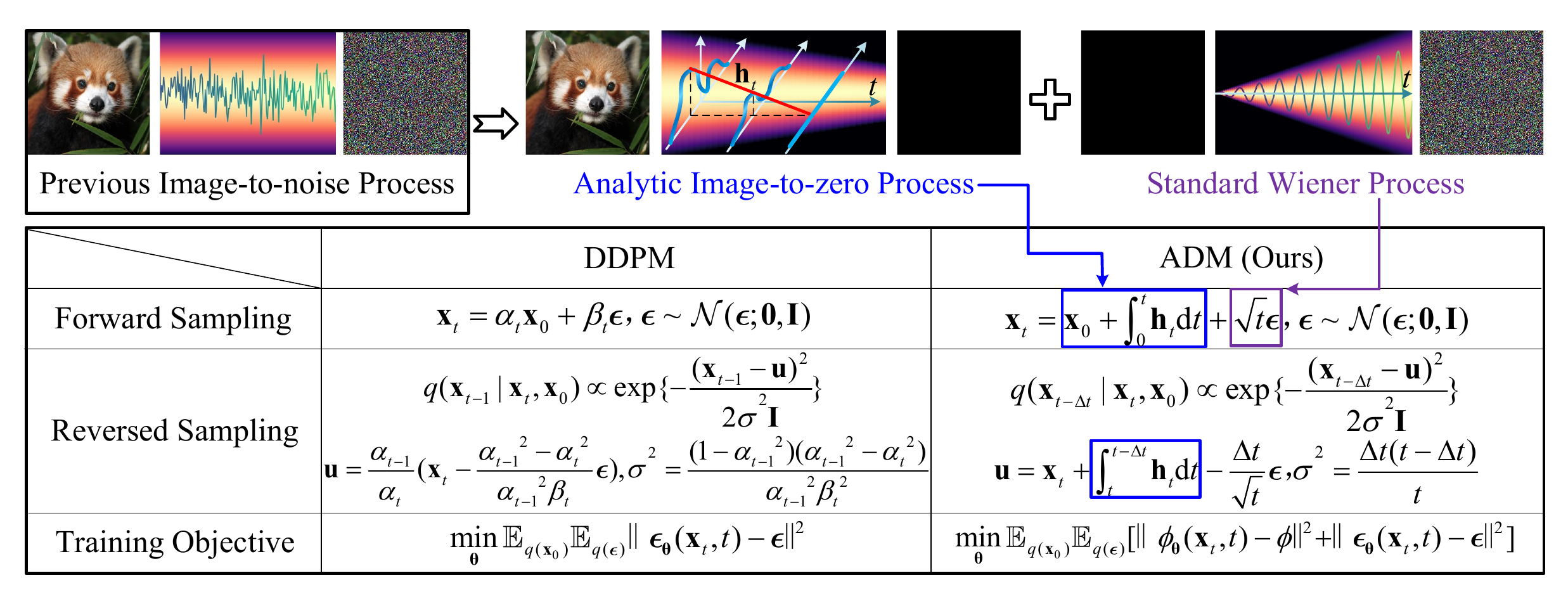}
    % \put(88, 26.3){\small $t$}
    % \put(88, 9.2){\small $t$}
    % \put(74.0, 27.7){\small $\mathbf{h}_t$}
    % \put(21.0, 0){\small JPPNet\cite{(25)liang2018look}}
    % \put(41.5, 0){\small MuLA\cite{(26)nie2018mutual}}
  \end{overpic} \vspace{-10pt}
  \caption{Framework overview. (\textbf{Top}:) DPMs typically use an image-to-noise process, while we propose to replace it with two relatively simpler processes: the image-to-zero mapping and the zero-to-noise mapping. We use an \textit{analytic} function (in \textcolor{blue}{blue} boxes) to model image-to-zero, or the attenuation gradient of the image (\textcolor{red}{red} line in the middle image with a linear attenuation). The zero-to-noise path %we draw the uniform attenuation gradient $\mathbf{h}_{t}$ in the red curve as an example). 
  is governed by the standard Wiener process. (\textbf{Bottom}:) We compare the equations of forward sampling, reversed sampling, and training objectives of our method and DDPM. Note that $\mathbf{h}_{t}$ is the proposed analytical image attenuation function and $\boldsymbol{\phi}$ represents its hyperparameter.}% whose variance increases linearly over time, enabling the zero image to progressively increase to the standard normal noise.}
\label{fig:1}\vspace{-10pt}
\end{figure*} 

This paper proposes a novel forward diffusion process to improve the generative ability of DPMs. As shown in Fig.~\ref{fig:1}, instead of using the normal image-to-noise forward diffusion, we introduce the analytical image attenuation into the forward diffusion, resulting in simultaneous image-to-zero and zero-to-noise processes. Concretely, the image-to-zero process attenuates the image component gradually while the zero-to-noise process increases the noise component until it obeys the standard normal distribution. Importantly, we have designed various analytical attenuation functions to control the image-to-zero process, thereby incorporating the analytical solution of the image-to-zero into the reversed diffusion process. In this approach, we derive our training objective: instead of predicting only noise from $\mathbf{x}_t$, we predict the clean image and noise independently from $\mathbf{x}_t$. We derive the sampling formula based on the analytic attenuation function for reversed diffusion. We name our method \textbf{D}iffusion \textbf{M}odels with \textbf{A}nalytical Image Attenuation (ADM).

% decouple this process: attenuate the image component by an image-to-zero mapping and increase the noise component by a zero-to-noise mapping. To model the latter, we use the standard Wiener process \cite{(58)einstein1905motion}. To model the former, we design various analytic functions such as $\mathbf{h}_{t}=\mathbf{c}$. 
% In this approach, we derive our training objective: instead of predicting only noise from $\mathbf{x}_t$, we predict the clean image and noise independently from $\mathbf{x}_t$. We derive the sampling formula based on the analytic transition function for reversed diffusion. 

\begin{figure*}[ht]
  \centering
  %\fbox{\rule[-.5cm]{0cm}{4cm} \rule[-.5cm]{4cm}{0cm}}
  %\includegraphics[width=1\linewidth]{figures/framework.pdf}
  \begin{overpic}
    [width=1\textwidth]{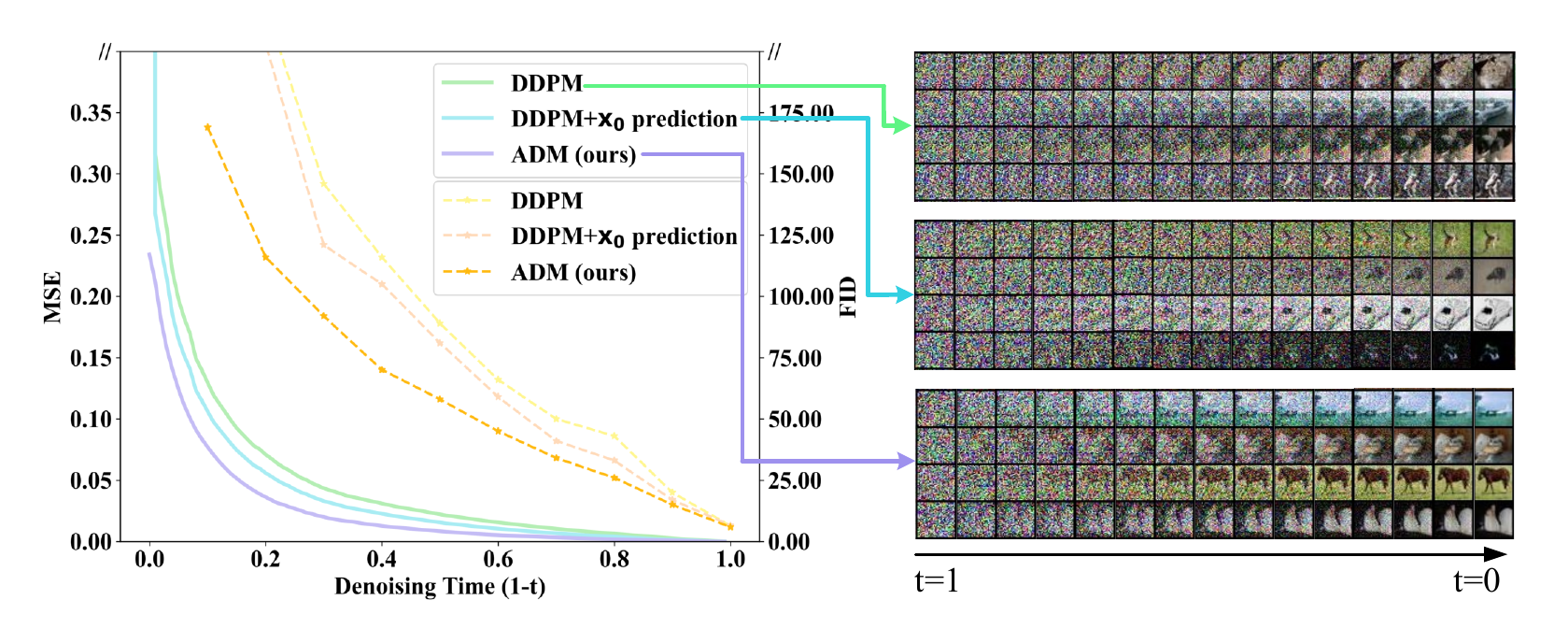}
    % \put(47, 29){\small DDPM}
    % \put(66, 29){\small DDPM+$\mathbf{x}_0$ prediction branch}
    % \put(85, 29){\small SDM (ours)}
  \end{overpic}%\vspace{-15pt}
  \caption{Comparing DDPM, its improved version, and ADM of image quality of unconditioned generation on CIFAR-10. (\textbf{Left}:) We evaluate image quality at each step using mean square error (MSE, solid line) and Fréchet inception distance (FID, dashed line) of the final $\mathbf{x}_{0}$ ($t=0$) and the estimated $\mathbf{\hat{x}}_{0}$ during denoising time. (\textbf{Right}:) Sample generated images from $t=0.4$ for the three compared methods. Improvement can be observed after adding to DDPM a branch that predicts $\mathbf{x}_0$. Further improvement is confirmed when the image-to-noise process is completely split to predict $\mathbf{x}_0$ and noise simultaneously.} %We plot the reversed diffusion 
  %Toy examples on CIFAR-10 dataset. We evaluate the efficacy of each step using the MSE of the final $\mathbf{x}_{0}$ and the estimated $\mathbf{\hat{x}}_{0}$ during denoising time. 
  %As shown in the figure, DDPM$^2$ predicts the noise and image components simultaneously, getting a better performance than DDPM, while DDPM$^3$ decouples the two parts more thoroughly, obtaining the best efficacy and visual result.}
  \label{fig:2}%\vspace{-10pt}
\end{figure*}

ADM has two benefits. First, by representing the image-to-noise mapping as two relatively simpler processes, we empirically find it improves the effectiveness of diffusion model training prominently. % reduce the learning difficulty of the model. 
In Fig.~\ref{fig:2}, either adding a branch to predict the image component or decoupling forward diffusion into two branches allows the denoising process to achieve high quality with much fewer evaluations. 
%On the other hand, 
%As shown in Fig.~\ref{}, we decouple the original forward equation of DDPM \cite{} into two simpler sub-process, enabling more precise and efficient solutions to the reversed process.
Second, during reversed diffusion, the image component can be obtained by an \textit{analytical} solution, which allows the reversed-time SDE to be solved with an arbitrary step size, thus significantly reducing the number of function evaluations. This benefit is ensured by the use of an \textit{analytic} attenuation function for modeling image-to-zero mapping. This offers a significant advantage over typical DPMs, where solving the reverse-time process accurately requires numerical integration, resulting in thousands of function evaluations. 

%\textbf{Why DDM allows for fewer function evaluations?}
%Previous DPMs directly model the image-to-noise process, which requires numerical integration with very small step size since it is not analytic. Differently, the DDM has an analytic image-to-zero process, allowing the reversed process to be solved with arbitrary step sizes, so we can generate samples with fewer function evaluations.
We summarize the main points of this paper below. %diffusion schemes,  %during reversed diffusion, this mapping can be predicted, and thus the image component can be obtained by an \textit{analytical} solution. 

\begin{itemize}[leftmargin=*]
\item We introduce a novel diffusion process ADM, %a new diffusion method named DDM, 
where the image-to-noise process is represented as simultaneous image-to-zero and zero-to-noise processes. %component and the image component from the diffusion process. 

\item We derive the training objective and analytical forward/reversed SDE. Via the former, ADMs learn to predict the image and noise simultaneously from $\mathbf{x}_{t}$. % image-to-zero mapping and zero-to-noise mapping. 
Via the latter, we avoid time-consuming numerical integration and achieve high-quality image generation with much fewer steps. %DDMs learn to predict image-to-zero mapping and zero-to-noise mapping. Through the latter, A novel forward diffusion SDE and its training objective that enables the model to learn the image and noise components respectively. Additionally, we derive the reverse denoising formula of DDM that naturally supports fewer steps of generation.

\item ADMs yield competitive performance in unconditional image generation under few function evaluation setups. Moreover, ADMs can be easily applied to downstream image-conditioned applications such as image super-resolution, saliency detection, edge detection, and image inpainting, and give very competitive accuracy with no more than 10 function evaluations. %Extensive experiments on various tasks including unconditional and conditional generation demonstrate the significant advantages.
\end{itemize}

\section{Related Work}
\textbf{Diffusion probabilistic models} 
%Diffusion probabilistic models %are an emerging generative paradigm that 
have demonstrated outstanding capabilities in generating images \cite{(8)dhariwal2021diffusion, (61)10021824, (63)10251995}, graph \cite{(62)10366850}, speech and music \cite{(29)DBLP:conf/ismir/MittalEHS21}, and 3D shapes \cite{(30)muller2022diffrf}. %As the proposed DDM provides an alternative framework for common tasks, we primarily focus on discussing the associated diffusion paradigms. 
\cite{(10)sohl2015deep} formulated diffusion probabilistic models, and \cite{(2)ddpm} first proposed a general diffusion framework for computer vision applications. \cite{(9)sde} established a link between DPMs and score-based generative models and derived general SDE frameworks for both the forward and reversed processes. Later, \cite{(23)DBLP:conf/cvpr/RombachBLEO22} and \cite{(27)DBLP:conf/nips/VahdatKK21} translated the diffusion process from the image space to latent space with an auto-encoder, allowing the model to generate higher-resolution images. Their attempts mainly translate the data space into a smoother one instead of improving the diffusion process itself. For the latter, \cite{(25)DBLP:conf/iclr/DockhornVK22} introduce velocity variables for the diffusion process for better performance and saving sampling time. They couple velocity with the image space, resulting in a more complex diffusion process. In comparison, we propose to split the diffusion process into two relatively simpler processes, where an analytic function is used for modeling the image-to-zero process, allowing for faster reversed diffusion. % introducing an analytic to model the gradient of the image component. % 添加 Flow one-step

\textbf{Few-step generation with DPMs} aims to %Recently, some researchers have focused on 
accelerate the sampling process. %There are two main methods that have been developed to achieve this. 
One line of works \cite{(33)DBLP:conf/iclr/SongME21,(34)DBLP:conf/nips/0011ZB0L022,(35)DBLP:conf/nips/FaradonbehFB22,(36)DBLP:conf/nips/DoucetGMS22,(37)DBLP:conf/nips/DockhornVK22, (38)zhang2022fast} construct ODE solvers of one or higher order to accelerate numerical integration in the reversed process. Essentially, they do not change the diffusion paradigm. %but rather correct the reversed process. 
A potential downside is that samples generated from these ODE solvers are determined by the initial noise thus exhibiting less diversity. In the other line of works, distillation-based methods \cite{(39)salimans2022progressive, (40)song2023consistency, (41)meng2022distillation, (42)luhman2021knowledge} use a fitted teacher model to distill the student model, allowing the latter to generate with few steps. In these methods, training is usually divided into a few stages, which increases the training cost. Differently, this paper does not focus on improving the sampling algorithms or training skills: the proposed ADM is an enhanced diffusion method that naturally supports generating high-quality images with fewer steps.

%\vspace{-5pt}
\section{Preliminaries}
The forward diffusion process of a typical DPM \cite{(10)sohl2015deep} is described as a Markov chain. Consider a continuous-time Markov chain with $t\in [0, 1]$:
\begin{equation}
 q(\mathbf{x}_{t}|\mathbf{x}_{0}) = \mathcal{N}(\mathbf{x}_{t}; \alpha_{t}\mathbf{x}_{0}, \beta_{t}^{2}\mathbf{I}),
 \label{1}
\end{equation}
where $\alpha_{t}, \beta_{t}$ are differentiable functions of time $t$ with bounded derivatives. $\beta_{t}$ is designed to increase gradually over time while $\alpha_{t}$ does the opposite, ensuring $q(\mathbf{x}_{1}|\mathbf{x}_{0}) = \mathcal{N}(\mathbf{x}_{1}; \mathbf{0}, \mathbf{I})$. It is proven that this Markov chain can be represented by the following SDE \cite{(12)kingma2021variational}:
\begin{equation}
 \mathrm{d}\mathbf{x}_{t} = f_{t}\mathbf{x}_{t}\mathrm{d}t + g_{t}\mathrm{d}\mathbf{w}_{t},\quad\mathbf{x}_{0} \sim q(\mathbf{x}_{0}),
 \label{2}
\end{equation}
where $f_{t}=\frac{\mathrm{d}\log{\alpha_{t}}}{\mathrm{d}t}$, $g_{t}^{2}=\frac{\mathrm{d}\beta_{t}^{2}}{\mathrm{d}t}-2f_{t}\beta_{t}^{2}$, and $\mathbf{w}_{t}$ is the standard Wiener process. \cite{(9)sde} shows that the corresponding reversed SDE inverting $\mathbf{x}_{1}$ to $\mathbf{x}_{0}$ can be derived as: 
\begin{equation}
 \mathrm{d}\mathbf{x}_{t} = [f_{t}\mathbf{x}_{t} - g_{t}^{2}\nabla_{\mathbf{x}}\log{q(\mathbf{x}_{t})}]\mathrm{d}t + g_{t}\mathrm{d}\overline{\mathbf{w}}_{t},
 \label{3}
\end{equation}
where $\overline{\mathbf{w}}_{t}$ is a standard Wiener process in the reversed diffusion. The only unknown term in Eq.~\ref{3},  $\nabla_{\mathbf{x}}\log{q(\mathbf{x}_{t})}$, is given by a neural network $\boldsymbol{\epsilon}_{\boldsymbol{\theta}}(\mathbf{x}_{t}, t)$ parameterized by $\boldsymbol{\theta}$. $\boldsymbol{\epsilon}_{\boldsymbol{\theta}}(\mathbf{x}_{t}, t)$ estimates the scaled score function, \textit{i.e.}, $-\beta_{t}\nabla_{\mathbf{x}}\log{q(\mathbf{x}_{t})}$, and the training objective \cite{(2)ddpm} is:
\begin{equation}
% \setlength{\abovedisplayskip}{3pt}
% \setlength{\belowdisplayskip}{3pt}
% \resizebox{\linewidth}{!}{
\begin{aligned}
 \mathcal{L}(\boldsymbol{\theta}, \lambda_{t}) &:=\int_{0}^{1} \lambda_{t} \mathbb{E}_{q(\mathbf{x}_{0})} \mathbb{E}_{q(\boldsymbol{\epsilon})}[||\boldsymbol{\epsilon}_{\boldsymbol{\theta}}(\mathbf{x}_{t}, t) \\&+ \beta_{t}\nabla_{\mathbf{x}}\log{q(\mathbf{x}_{t})}||^{2}] \mathrm{d}t \\
      &=\int_{0}^{1} { \lambda_{t} \mathbb{E}_{q(\mathbf{x}_{0})} \mathbb{E}_{q(\boldsymbol{\epsilon})} [\Vert{\boldsymbol{\epsilon}_{\boldsymbol{\theta}}(\mathbf{x}_{t}, t)-\boldsymbol{\epsilon}}\Vert^{2}] \mathrm{d}t},
 \end{aligned}
 % }
 \label{4}
\end{equation}
where $\boldsymbol{\epsilon}\sim \mathcal{N}(\boldsymbol{\epsilon}; \mathbf{0}, \mathbf{I}), \mathbf{x}_{t}=\alpha_{t}\mathbf{x}_{0}+\beta_{t}\boldsymbol{\epsilon}$, and $\lambda_{t}$ is a weighting function. 
    
Replacing the score function by $\boldsymbol{\epsilon}_{\boldsymbol{\theta}}(\mathbf{x}_{t}, t)$ in Eq.~\ref{3}, one can generate samples by solving the following reversed SDE with numerical integration, starting from $\mathbf{x}_{1}\sim \mathcal{N}(\mathbf{x}_{1}; \mathbf{0}, \mathbf{I})$:
\begin{equation}
 \mathrm{d}\mathbf{x}_{t} = [f_{t}\mathbf{x}_{t} + \frac{g_{t}^{2}}{\beta_{t}}\boldsymbol{\epsilon}_{\boldsymbol{\theta}}(\mathbf{x}_{t}, t)]\mathrm{d}t + g_{t}\mathrm{d}\overline{\mathbf{w}}_{t}.
 \label{5}
\end{equation}
Due to the complexity of the SDE, it is difficult to estimate the score function accurately and let alone finding an analytical solution for the reverse-time SDE. %and to find an analytical solution for the reverse-time SDE. 
As a result, DPMs need thousands of function evaluations to generate high-quality images, resulting in extremely long inference time.
    % \begin{equation}
    % \footnotesize
    % E(\mathcal{A}) =
    % \sum_{\alpha_{i,j}\in\mathcal{A}}
    % {-
    % \frac{\exp{(\alpha^{o}_{i,j})}} {\sum_{o^{'}\in\mathcal{O}} \exp{(\alpha^{o^{'}}_{i,j})}}
    % }\cdot\log(\frac{\exp{(\alpha^{o}_{i,j})}} {\sum_{o^{'}\in\mathcal{O}} \exp{(\alpha^{o^{'}}_{i,j})}})
    % \label{9}
    % \end{equation}
\section{Method}
% \begin{figure*}
%   \centering
%   %\fbox{\rule[-.5cm]{0cm}{4cm} \rule[-.5cm]{4cm}{0cm}}
%   %\includegraphics[width=1\linewidth]{figures/framework.pdf}
%   \begin{overpic}
%     [width=1\textwidth]{figures/preliminary-cvpr.pdf}
%     % \put(47, 29){\small DDPM}
%     % \put(66, 29){\small DDPM+$\mathbf{x}_0$ prediction branch}
%     % \put(85, 29){\small SDM (ours)}
%   \end{overpic}\vspace{-15pt}
%   \caption{Comparing DDPM, its improved version, and SDM of image quality of unconditioned generation on CIFAR-10. (\textbf{Left}:) We evaluate image quality at each step using mean square error (MSE, solid line) and Fréchet inception distance (FID, dashed line) of the final $\mathbf{x}_{0}$ ($t=0$) and the estimated $\mathbf{\hat{x}}_{0}$ during denoising time. (\textbf{Right}:) Sample generated images from $t=0.4$ for the three compared methods. Improvement can be observed after adding to DDPM a branch that predicts $\mathbf{x}_0$. Further improvement is confirmed when the image-to-noise process is completely split to predict $\mathbf{x}_0$ and noise simultaneously.} %We plot the reversed diffusion 
%   %Toy examples on CIFAR-10 dataset. We evaluate the efficacy of each step using the MSE of the final $\mathbf{x}_{0}$ and the estimated $\mathbf{\hat{x}}_{0}$ during denoising time. 
%   %As shown in the figure, DDPM$^2$ predicts the noise and image components simultaneously, getting a better performance than DDPM, while DDPM$^3$ decouples the two parts more thoroughly, obtaining the best efficacy and visual result.}
%   \label{fig:2}\vspace{-10pt}
% \end{figure*}

\begin{algorithm}[t]  
  \caption{Training algorithm of ADM.}  
  \begin{algorithmic}[1]  
    % \Require  
    %   The proportion of the operation $o(\cdot)$ in connection from node $j$ to node $i$, $\gamma^{o}_{i,j}$, $j\leq i$;
    % \Ensure  
    %   The remaining operations list, $remain_o$;  
        \State \textbf{Initialize} $i=0, N=num\_iters, lr,$; network parameters: $\boldsymbol{\theta}$;%of $\mathrm{Net}_{\boldsymbol{\theta}}$;
        %\label{code:fram:extract}  
        \While {$i<N$}
            \State $t\sim Uniform(0, 1), \mathbf{x}_{0}\sim q(\mathbf{x}_{0}), \boldsymbol{\epsilon}\sim \mathcal{N}(\mathbf{0}, \mathbf{I}), \boldsymbol{\phi}\sim \mathbf{x}_{0} + \mathbf{H}_{1}=\mathbf{0}$;
            \State $\mathbf{H}_{t} = \int_{0}^{t} {\mathbf{h}^{\boldsymbol{\phi}}_{t} \mathrm{d}t}$;
            \State $\mathbf{x}_{t} = \mathbf{x}_{0} + \mathbf{H}_{t} + \sqrt{t}\boldsymbol{\epsilon}$;
            \State $\boldsymbol{\phi}_{\boldsymbol{\theta}}, \boldsymbol{\epsilon}_{\boldsymbol{\theta}} = \mathrm{Net}_{\boldsymbol{\theta}}(\mathbf{x}_{t}, t)$;
            \State $\boldsymbol{\theta} \leftarrow lr*\nabla_{\boldsymbol{\theta}} (\Vert \boldsymbol{\phi}_{\boldsymbol{\theta}}-\boldsymbol{\phi}\Vert^{2} + \Vert \boldsymbol{\epsilon}_{\boldsymbol{\theta}}-\boldsymbol{\epsilon}\Vert^{2})$;
            \State $i = i+1$;
        \EndWhile; \\
        \Return $\boldsymbol{\theta}$;  
    \end{algorithmic}  
    \label{alg:1}
\end{algorithm}

\subsection{Preliminary Experiments: Improving DDPM with \textbf{x}$_{0}$ Prediction} \label{sec:4.1}

To show the benefit of incorporating the analytical image attenuation into the forward diffusion, we use DDPM \cite{(2)ddpm} as baseline and create an improved version. %, and compare both with the proposed SDM. 
The two methods are described below. %gradually make changes to it.  Three diffusion methods are compared. % to train the network. 

\begin{itemize}[leftmargin=*]
    \item DDPM. It has a small U-Net architecture ($48$M parameters) and can be trained using only the noise loss \cite{(2)ddpm}. Because this paper uses continuous time $t\in [0, 1]$, we linearly transform the discrete time $t=1, 2,...,1000$ used in DDPM into $t\in [0, 1]$ with interval 0.001. %Normally DDPM uses discrete time , but , we we align it with the continuous time $t\in [0, 1]$ that each step is equivalent to $0.001$.
    \item DDPM+$\mathbf{x}_0$ prediction. On top of DDPM, we add a convolutional branch to predict $\mathbf{x}_{0}$. In practice, we use a small U-Net architecture with the feature channel 64 and channel multiplier [1, 2, 4, 8]. We train all preliminary models for only 200k iterations.%Details of this branch are provided in Supplementary Materials. 
    %\item SDM. We use the same U-Net structure and noise loss as ``DDPM+$x_0$ prediction'' but completely decouple the mapping from image to noise into two simpler parts: an image-to-zero process and a zero-to-noise process.  %diffusion into the proposed two terms: a drift term that gradually degrades the original image to the zero image, and a diffusion term that transforms the zero image to standard normal noise along the time. 
\end{itemize}

In Fig. \ref{fig:2}, we compare the unconditioned image generation quality of the three methods. Specifically, after obtaining an estimation of $\mathbf{x}_{0}$ %by the forward transition probability $q(\mathbf{x}_{t}|\mathbf{x}_{0})$ 
at any time $t$ in the reverse-time process, we use the mean squared error (MSE) between the estimated $\mathbf{\hat{x}}_{0}$ and $\mathbf{x}_{0}$ to indicate image quality generated at time $t$. 

%Toy Examples}
%To explore how to improve the effectiveness of DPMs, we construct a series of variants to compare the efficacy of each reversed step. As we can get an approximate estimation of $\mathbf{x}_{0}$ by the forward transition probability $q(\mathbf{x}_{t}|\mathbf{x}_{0})$ for any time $t$ in the reverse-time process, we use the mean squared error (MSE) between the estimated $\mathbf{\hat{x}}_{0}$ and $\mathbf{x}_{0}$ to evaluate the efficacy of time $t$. 
    
%Here we take DDPM \cite{(2)ddpm} with a small U-net architecture ($48$M parameters) as our baseline and only use the noise loss \cite{(2)ddpm} to train the network. Since DDPM utilizes the discrete time from $1$ to $1000$, we align it with the continuous time $t\in [0, 1]$ that each step is equivalent to $0.001$. 

From Fig.~\ref{fig:2}, at $t\approx0.5$, images generated by the vanilla DDPM are quite close to noise. In fact, Fig. \ref{fig:2} shows that it usually takes $900\sim1000$ steps ($t=0.1\sim0$) for DDPM to generate meaningful images. In comparison, quality of images generated by ``DDPM+$\mathbf{x}_0$ prediction'' is much higher than the vanilla DDPM. It suggests %1) that predicting only noise from $x_t$ is a difficult task and 2) 
that predicting both noise and $\mathbf{x}_0$ from $\mathbf{x}_t$ with additional $\mathbf{x}_0$ supervision benefits diffusion model training.

While these experiments indicate the usefulness of $\mathbf{x}_0$ prediction, in DDPM, $\mathbf{x}_0$ is still deeply mixed with noise in $\mathbf{x}_t$. This consideration motivates us to explicitly and separately model $\mathbf{x}_0$ along time. In fact, using the proposed $\mathbf{x}_0$ modeling method ADM with the same U-Net structure and noise loss as ``DDPM+$\mathbf{x}_0$ prediction'', image generation quality is further improved (see Fig.~\ref{fig:2}).\footnote{For ADM model used in the preliminary experiments, we add an additional decoder based on the small U-Net and use the constant attenuation function that $\mathbf{h}_{t}=\mathbf{c}$.}

\subsection{Diffusion Models with Analytical Image Attenuation}\label{sec:4.2}% with Explicit Transition Probability}
%The preliminary experiments provide guidance for improving the generative abilities of DPMs, however, it cannot handle the numerical integration that require thousands of function evaluations when solving the reverse-time SDE. From another perspective, the curves in Fig~.\ref{fig:2} can be viewed as the variation of $\mathbf{x}_{t}$ with respect to $t$; that is, if we obtain the explicit expression of $\frac{\mathrm{d}\mathbf{x}_{t}}{\mathrm{d}t}$, the reversed process can be expedited by the analytic solution.  Building on this insight, our SDM models the gradient of the image component as an explicit and differentiable function, offering the advantages of high efficacy and high speed.

%SDM decouples the image-to-noise process into image-to-zero and zero-to-noise processes. 
This section introduces a forward process and a reversed process, representing the pathways of image corruption and denoising, respectively. We prove that the proposed forward diffusion process is equivalent to the previous mapping from image to noise. % based on the fact they describe the same mapping from image to noise. 
%To demonstrate its effectiveness, 
We then derive the training objective of ADM. % by maximizing the evidence lower bound of the log-likelihood. 
Finally, we use the Bayesian theorem to derive the sampling formula of ADM which, due to its analytic nature, allows for sampling with arbitrary step sizes and thus few-step generation.

\begin{algorithm}[t]
  \caption{Sampling algorithm of ADM.}  
  \begin{algorithmic}[1]  
    % \Require  
    %   The proportion of the operation $o(\cdot)$ in connection from node $j$ to node $i$, $\gamma^{o}_{i,j}$, $j\leq i$;
    % \Ensure  
    %   The remaining operations list, $remain_o$;  
    \State \textbf{Initialize} $\mathbf{x}_{t}\sim \mathcal{N}(\mathbf{0}, \mathbf{I}),t=1,$ \;
        $N=num\_steps, s=1/N, \mathrm{Net}_{\boldsymbol{\theta}}$;  
    %\label{code:fram:extract}  
    \While {$t>0$}
        \State $\boldsymbol{\phi}_{\boldsymbol{\theta}}, \boldsymbol{\epsilon}_{\boldsymbol{\theta}} = \mathrm{Net}_{\boldsymbol{\theta}}(\mathbf{x}_{t}, t)$;
        \State $\mathbf{H}_{t} = \int_{0}^{t} {\mathbf{h}^{\boldsymbol{\phi}}_{t} \mathrm{d}t}$;
        \State $\mathbf{\widetilde{u}} = \mathbf{x}_{t} + \mathbf{H}_{t-s}-\mathbf{H}_{t}-\frac{s}{\sqrt{t}}\boldsymbol{\epsilon}_{\boldsymbol{\theta}}$;
        \State $\widetilde{\sigma} = \sqrt{\frac{s(t-s)}{t}}$;
        \State $\mathbf{x}_{t} = \mathbf{\widetilde{u}} + \widetilde{\sigma}*\boldsymbol{\widetilde{\epsilon}}$,\ \ $\boldsymbol{\widetilde{\epsilon}}\sim \mathcal{N}(\mathbf{0}, \mathbf{I})$;
        \State $t = t-s$;
    \EndWhile; \\
    \Return $\mathbf{x}_{t}$;  
  \end{algorithmic}  
  \label{alg:2}
\end{algorithm}
% \begin{algorithm}[t]  
%   \caption{Training algorithm of ADM.}  
%   \begin{algorithmic}[1]  
%     % \Require  
%     %   The proportion of the operation $o(\cdot)$ in connection from node $j$ to node $i$, $\gamma^{o}_{i,j}$, $j\leq i$;
%     % \Ensure  
%     %   The remaining operations list, $remain_o$;  
%         \State \textbf{Initialize} $i=0, N=num\_iters, lr,$; network parameters: $\boldsymbol{\theta}$;%of $\mathrm{Net}_{\boldsymbol{\theta}}$;
%         %\label{code:fram:extract}  
%         \While {$i<N$}
%             \State $t\sim Uniform(0, 1), \mathbf{x}_{0}\sim q(\mathbf{x}_{0}), \boldsymbol{\epsilon}\sim \mathcal{N}(\mathbf{0}, \mathbf{I}), \boldsymbol{\phi}\sim \mathbf{x}_{0} + \mathbf{H}_{1}=\mathbf{0}$;
%             \State $\mathbf{H}_{t} = \int_{0}^{t} {\mathbf{h}^{\boldsymbol{\phi}}_{t} \mathrm{d}t}$;
%             \State $\mathbf{x}_{t} = \mathbf{x}_{0} + \mathbf{H}_{t} + \sqrt{t}\boldsymbol{\epsilon}$;
%             \State $\boldsymbol{\phi}_{\boldsymbol{\theta}}, \boldsymbol{\epsilon}_{\boldsymbol{\theta}} = \mathrm{Net}_{\boldsymbol{\theta}}(\mathbf{x}_{t}, t)$;
%             \State $\boldsymbol{\theta} \leftarrow lr*\nabla_{\boldsymbol{\theta}} (\Vert \boldsymbol{\phi}_{\boldsymbol{\theta}}-\boldsymbol{\phi}\Vert^{2} + \Vert \boldsymbol{\epsilon}_{\boldsymbol{\theta}}-\boldsymbol{\epsilon}\Vert^{2})$;
%             \State $i = i+1$;
%         \EndWhile; \\
%         \Return $\boldsymbol{\theta}$;  
%     \end{algorithmic}  
%     \label{alg:1}
% \end{algorithm}

\textbf{Forward process}. We propose a novel diffusion formula %based on the It\^{o} process 
to describe the forward process:
\begin{equation}
 \mathbf{x}_{t} = \mathbf{x}_{0} + \int_{0}^{t} {\mathbf{h}_{t}\mathrm{d}t} + \int_{0}^{t} {\mathrm{d}\mathbf{w}_{t}},
 \quad\mathbf{x}_{0} \sim q(\mathbf{x}_{0}).
 \label{7}
\end{equation}
Here, $\mathbf{x}_{0} + \int_{0}^{t} {\mathbf{h}_{t}\mathrm{d}t}$ describes the image attenuation process, \emph{i.e.}, image to zero, and $\int_{0}^{t} {\mathrm{d}\mathbf{w}_{t}}$ denotes the noise increasing process, \emph{i.e.}, zero to noise. $\mathbf{h}_{t}$ is a differentiable function of $t$, and $\mathbf{w}_{t}$ is the standard Wiener process. Eq.~\ref{7} shows that $q(\mathbf{x}_{t}|\mathbf{x}_{0})$ can be regarded as a normal distribution with mean $\mathbf{x}_{0} + \int_{0}^{t} {\mathbf{h}_{t}\mathrm{d}t}$ and variance $t\mathbf{I}$. We prove that the proposed diffusion process defined in Eq.~\ref{7} is equivalent to the previous SDE in Section I-A of Supplementary Material. %Appendix~\ref{supsec:1.1}.%Section 1.1 of Supplementary Material. 

The forward process must ensure that $\mathbf{x}_{t}\sim q(\mathbf{x}_{0})$ when $t=0$ and $\mathbf{x}_{t}\sim \mathcal{N}(\mathbf{x}_{1}; \mathbf{0}, \mathbf{I})$ when $t=1$. Obviously, Eq.~\ref{7} satisfies the first requirement when setting $t=0$. Moreover, the noise term in Eq.~\ref{7} itself meets the second requirement because $\int_{0}^{t} {\mathrm{d}\mathbf{w}_{t}} = \mathbf{w}_{t} - \mathbf{w}_{0} \sim \mathcal{N}(\mathbf{0}, t\mathbf{I})$. %Meanwhile, the image attenuation process $\mathbf{x}_{0} + \int_{0}^{t} {\mathbf{h}_{t}\mathrm{d}t}$ satisfies  
So, we only need to design $\mathbf{h}_{t}$ such that: $\mathbf{x}_{0} + \int_{0}^{1} {\mathbf{h}_{t}\mathrm{d}t}=\mathbf{0}$. Note that \textit{$\mathbf{h}_{t}$ can be analytic, which will significantly improve the efficiency of the reversed process.} We analyze various forms of $\mathbf{h}_{t}$ in Section~\ref{sec:exp}.%A linear form $\mathbf{h}_{t} = \mathbf{c}$ was used in Fig. \ref{fig:2} and various choices of $\mathbf{h}_{t}$ will be compared in Section \ref{sec:exp}, . %\textcolor{blue}{A linear form $\mathbf{h}_{t} = c$ where $c$ is a learnable parameter was used in Fig. \ref{fig:2}. In Section \ref{sec:exp}, various choices of $\mathbf{h}_{t}$ will be compared.}

For simplicity, we denote $\mathbf{H}_{t} = \int_{0}^{t} {\mathbf{h}_{t}\mathrm{d}t}$ and give the distribution of $\mathbf{x}_{t}$ conditioned on $\mathbf{x}_{0}$:
\begin{equation}
 q(\mathbf{x}_{t}|\mathbf{x}_{0}) = \mathcal{N}(\mathbf{x}_{t}; \mathbf{x}_{0}+\mathbf{H}_{t}, t\mathbf{I}).
 \label{8}
\end{equation}
Thus, we can sample $\mathbf{x}_{t}$ by $\mathbf{x}_{t}=\mathbf{x}_{0}+\mathbf{H}_{t}+\sqrt{t}\boldsymbol{\epsilon}$ in the forward process, where $\boldsymbol{\epsilon}\sim \mathcal{N}(\mathbf{0}, \mathbf{I})$.

\textbf{Reversed process}. Based on the the distribution of $\mathbf{x}_{t}$ conditioned on $\mathbf{x}_{0}$, we build the reversed diffusion process using the Bayesian theorem. Unlike DDPM \cite{(2)ddpm} which uses \textit{discrete}-time Markov chain, we employ \textit{continuous}-time Markov chain with the smallest time step $\Delta t\rightarrow 0^{+}$ and use conditional distribution $q(\mathbf{x}_{t-\Delta t}|\mathbf{x}_{t}, \mathbf{x}_{0})$ to approximate $q(\mathbf{x}_{t-\Delta t}|\mathbf{x}_{t})$:
\begin{equation}
 q(\mathbf{x}_{t-\Delta t}|\mathbf{x}_{t}, \mathbf{x}_{0}) = \frac{q(\mathbf{x}_{t}|\mathbf{x}_{t-\Delta t}, \mathbf{x}_{0})q(\mathbf{x}_{t-\Delta t}|\mathbf{x}_{0})} {q(\mathbf{x}_{t}|\mathbf{x}_{0})}.
 \label{9}
\end{equation}
Given the forward diffusion formula, we derive the reversed transition probability $q(\mathbf{x}_{t-\Delta t}|\mathbf{x}_{t}, \mathbf{x}_{0})$ also follows a normal distribution:
\begin{equation}
\begin{split}
 &q(\mathbf{x}_{t-\Delta t}|\mathbf{x}_{t}, \mathbf{x}_{0}) \propto \exp\{-\frac{(\mathbf{x}_{t-\Delta t}-\mathbf{\widetilde{u}})^{2}}{2\widetilde{\sigma}^{2}\mathbf{I}}\}, \\
 &\mathbf{\widetilde{u}} = \mathbf{x}_{t}+\mathbf{H}_{t-\Delta t}-\mathbf{H}_{t}-\frac{\Delta t}{\sqrt{t}}\boldsymbol{\epsilon},\\
 &\widetilde{\sigma}^{2}=\frac{\Delta t(t-\Delta t)}{t},
 \end{split}
 \label{10}
\end{equation}
%where $\boldsymbol{\epsilon}\sim \mathcal{N}(\mathbf{0}, \mathbf{I})$.
In Eq. \ref{10}, the variance term $\widetilde{\sigma}^{2}$ is non-parametric and solely dependent on the current time $t$ and the step size $\Delta t$, whereas the mean term $\mathbf{\widetilde{u}}$ involves two unknown variables: $\mathbf{H}_{t}$ and $\boldsymbol{\epsilon}$ in the reversed process. As a result, to solve the reversed process, we must use $p_{\boldsymbol{\theta}}(\mathbf{x}_{t-\Delta t}|\mathbf{x}_{t})$ to approximate $q(\mathbf{x}_{t-\Delta t}|\mathbf{x}_{t}, \mathbf{x}_{0})$ and estimate $\mathbf{H}_{t}$ and $\boldsymbol{\epsilon}$ using a neural network. 
Essentially, the value of $\mathbf{H}_{t}$ depends on $\mathbf{h}_{t}$, an analytic function with respect to $t$, which % $\mathbf{h}_{t}$ 
is determined by its parameters $\boldsymbol{\phi}$. The ground truth of $\boldsymbol{\phi}$ can be obtained by solving $\mathbf{x}_{0} + \int_{0}^{1} {\mathbf{h}_{t}\mathrm{d}t}=\mathbf{0}$ and details can be found in Section~\ref{sec:exp.1}. %Section 2.3 of Supplementary Material. 
In training, we predict $\boldsymbol{\phi}$ and $\boldsymbol{\epsilon}$ by the U-Net architecture with two decoder branches: %such that they are obtained as the outputs of the network: 
$\boldsymbol{\phi}_{\boldsymbol{\theta}}, \boldsymbol{\epsilon}_{\boldsymbol{\theta}}=\mathrm{Net}_{\boldsymbol{\theta}}(\mathbf{x}_{t}, t)$.
%We refer readers to Section 1 and Section 2 of Supplementary Material for more details about derivations and U-Net architectures.

\begin{table*}[ht]
    \centering
    \caption{Hyper-parameters for the trained ADMs.} \vspace{3pt}
    \label{subtab:1}
    \tabcolsep=10pt
	\renewcommand\arraystretch{1.5}
    \resizebox{\textwidth}{!}{
    \begin{tabular}{lcccccc}
    \hline
    \multirow{2}{*}{Task} & \multicolumn{2}{c}{Unconditional generation}            & \multirow{2}{*}{Inpainting} & \multirow{2}{*}{\begin{tabular}[c]{@{}c@{}}Super\\ Resolution\end{tabular}} & \multirow{2}{*}{\begin{tabular}[c]{@{}c@{}}Saliency\\ Detection\end{tabular}} & \multirow{2}{*}{\begin{tabular}[c]{@{}c@{}}Edge\\ Detection\end{tabular}} \\ \cline{2-3}
                          & \multicolumn{1}{c}{CIFAR10}          & CelebA-HQ-256    &                             &                                                                             &                                                                               &                                                                           \\ \hline
    Image size            & \multicolumn{1}{c}{32$\times$32}  &  256$\times$256 &  256$\times$256   &  512$\times$512 &  384$\times$ 384   &    320$\times$320                                                                                                                   \\ \hline
    Batch size            & \multicolumn{1}{c}{128}              & 48               & 48                          & 12                                                                          & 16                                                                            & 16                                                                        \\ \hline
    Learning rate         & \multicolumn{1}{c}{1e-4$\sim$1e-5}    & 1e-4$\sim$1e-5 &   4e-5$\sim$4e-6   &  5e-5$\sim$5e-6    &             5e-5$\sim$5e-6      &     5e-5$\sim$5e-6                            \\ \hline
    Iterations            & \multicolumn{1}{c}{800k}             & 400k             & 400k                        & 200k                                                                        & 200k                                                                          & 200k                                                                      \\ \hline
    Feature channels      & \multicolumn{1}{c}{160}              & 128              & 96                          & 128                                                                         & 128                                                                           & 128                                                                       \\ \hline
    Channel multiplier    & \multicolumn{1}{c}{{[}1, 2, 2, 2{]}} & {[}1, 2, 2, 2{]} & {[}1, 2, 4, 8{]}            & {[}1, 2, 4, 4{]}                                                            & {[}1, 2, 4, 4{]}                                                              & {[}1, 2, 4, 4{]}                                                          \\ \hline
    Number of blocks      & \multicolumn{1}{c}{3}                & 3                & 2                           & 2                                                                           & 2                                                                             & 2                                                                         \\ \hline
    Smallest time step    & \multicolumn{1}{c}{1e-3}             & 1e-3             & 1e-4                        & 1e-4                                                                        & 1e-4                                                                          & 1e-4                                                                      \\ \hline
    \end{tabular}
    }
\end{table*}

% \begin{algorithm}[t]
%   \caption{Sampling algorithm of ADM.}  
%   \begin{algorithmic}[1]  
%     % \Require  
%     %   The proportion of the operation $o(\cdot)$ in connection from node $j$ to node $i$, $\gamma^{o}_{i,j}$, $j\leq i$;
%     % \Ensure  
%     %   The remaining operations list, $remain_o$;  
%     \State \textbf{Initialize} $\mathbf{x}_{t}\sim \mathcal{N}(\mathbf{0}, \mathbf{I}),t=1,$ \;
%         $N=num\_steps, s=1/N, \mathrm{Net}_{\boldsymbol{\theta}}$;  
%     %\label{code:fram:extract}  
%     \While {$t>0$}
%         \State $\boldsymbol{\phi}_{\boldsymbol{\theta}}, \boldsymbol{\epsilon}_{\boldsymbol{\theta}} = \mathrm{Net}_{\boldsymbol{\theta}}(\mathbf{x}_{t}, t)$;
%         \State $\mathbf{H}_{t} = \int_{0}^{t} {\mathbf{h}^{\boldsymbol{\phi}}_{t} \mathrm{d}t}$;
%         \State $\mathbf{\widetilde{u}} = \mathbf{x}_{t} + \mathbf{H}_{t-s}-\mathbf{H}_{t}-\frac{s}{\sqrt{t}}\boldsymbol{\epsilon}_{\boldsymbol{\theta}}$;
%         \State $\widetilde{\sigma} = \sqrt{\frac{s(t-s)}{t}}$;
%         \State $\mathbf{x}_{t} = \mathbf{\widetilde{u}} + \widetilde{\sigma}*\boldsymbol{\widetilde{\epsilon}}$,\ \ $\boldsymbol{\widetilde{\epsilon}}\sim \mathcal{N}(\mathbf{0}, \mathbf{I})$;
%         \State $t = t-s$;
%     \EndWhile; \\
%     \Return $\mathbf{x}_{t}$;  
%   \end{algorithmic}  
%   \label{alg:2}
% \end{algorithm}

\textbf{Training objective.} \label{sec:4.3}
By maximizing the evidence lower bound of the log-likelihood, %$\log{p(\mathbf{x})}$, 
we demonstrate the training objective of ADM (please refer to Section I-C of Supplementary Material):%Appendix~\ref{supsec:1.2} for details):
\begin{equation}
\min\limits_{\boldsymbol{\theta}} \mathbb{E}_{q(\mathbf{x}_{0})} \mathbb{E}_{q(\boldsymbol{\epsilon})} [\Vert \boldsymbol{\phi}_{\boldsymbol{\theta}}-\boldsymbol{\phi}\Vert^{2} + \Vert \boldsymbol{\epsilon}_{\boldsymbol{\theta}}-\boldsymbol{\epsilon}\Vert^{2}].
 \label{11}
\end{equation}
In Eq.~\ref{11}, the first term and second term correspond to the % in the training objective corresponds to the 
image component and noise component, respectively. %, while the subsequent term corresponds to the noise component. 
That is, the model is trained to predict image and noise separately from $\mathbf{x}_t$, in line with our process of simultaneous image-to-zero and zero-to-noise.%decoupling the clean image and noise components from $\mathbf{x}_{t}$. 
We attach two hyperparameters $\lambda_{1}$ and $\lambda_{2}$ to the first and second terms of Eq.~\ref{11}, respectively.
During the reverse-time process, %From an abstract viewpoint, 
the noise represents the starting point %in the reverse-time process 
for $\mathbf{x}_{t}$, whereas the clean image represents the endpoint. Simultaneously predicting the image and noise components allows $\mathbf{x}_{t}$ to be aligned with both starting and ending points at each time step. Detailed training procedure is presented in Alg.~\ref{alg:1}.

\begin{figure}[ht]
    \centering
    %\fbox{\rule[-.5cm]{0cm}{4cm} \rule[-.5cm]{4cm}{0cm}}
    %\includegraphics[width=1\linewidth]{figures/framework.pdf}
    \begin{overpic}
    [width=1\linewidth]{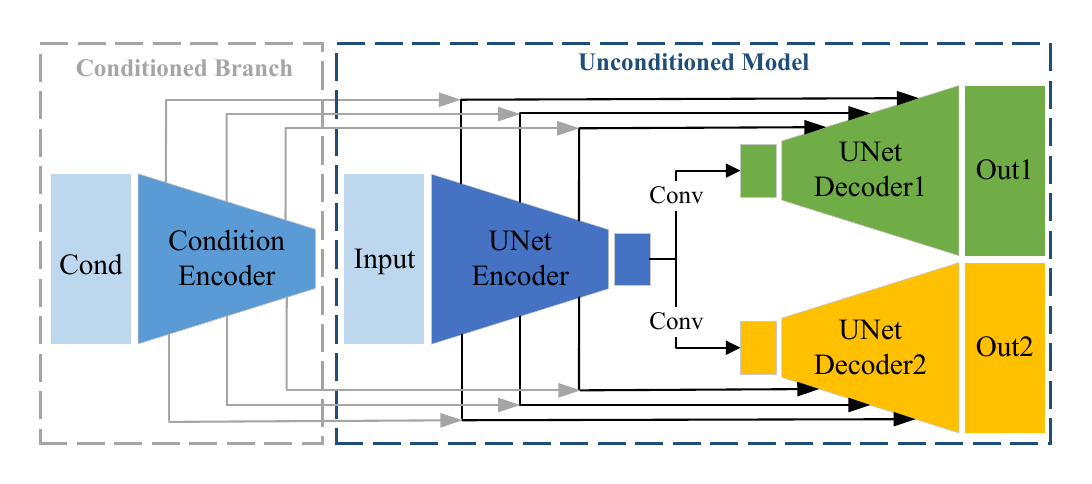}
    % \put(0.1, 21){\small Input}
    % \put(-0.5, 12){\small DDPM}
    % \put(0.2, 4.2){\small Ours}
    % \put(21.0, 0){\small JPPNet\cite{(25)liang2018look}}
    % \put(41.5, 0){\small MuLA\cite{(26)nie2018mutual}}
    \end{overpic} %\vspace{-15pt}
    \caption{Architecture detail. The architecture is built upon U-Net. Different from the vanilla U-Net, we construct two decoders to parameterize $\boldsymbol{\phi}$ and $\boldsymbol{\epsilon}$ simultaneously. The `Cond' represents the conditioned input and the conditional encoder is the general image backbone such as Swin \cite{(57)liu2021swin}. The conditional branch is only used in conditioned generation tasks.}\label{supfig:1}\vspace{-5pt}
\end{figure}

\textbf{Few-step sampling from ADM.} %\label{sec:4.4}
After introducing the reverse-time process in Section \ref{sec:4.2}, %preceding section. 
we now directly sample from ADM by iteratively solving Eq.~\ref{10} from $t=1$ to $0$. Because $\mathbf{h}_{t}$ has an analytic form, ADM possesses the natural property of sampling with an arbitrary step size $s$. For example, under linear image-to-zero assumption $\mathbf{h}_{t}=\mathbf{c}$, $\mathbf{H}_{t} = \int_{0}^{t} {\mathbf{h}_{t} \mathrm{d}t} = \mathbf{c}t$; as such, we can easily obtain $\mathbf{H}_{t-s} = \mathbf{c}(t-s)$. Consequently, we can derive the expression for $q(\mathbf{x}_{t-s}|\mathbf{x}_{t}, \mathbf{x}_{0})$ as:
\begin{equation}
\begin{split}
 &q(\mathbf{x}_{t-s}|\mathbf{x}_{t}, \mathbf{x}_{0}) \propto \exp\{-\frac{(\mathbf{x}_{t-s}-\mathbf{\widetilde{u}})^{2}}{2\widetilde{\sigma}^{2}\mathbf{I}}\}, \\
 &\mathbf{\widetilde{u}} = \mathbf{x}_{t}+\mathbf{H}_{t-s}-\mathbf{H}_{t}-\frac{s}{\sqrt{t}}\boldsymbol{\epsilon},\\
 &\widetilde{\sigma}^{2}=\frac{s(t-s)}{t}.
 \end{split}
 \label{12}
\end{equation}
Compared with the vanilla DDPM where the sampling step size has to be $s=0.001$ (aligning discrete time $0\sim 1000$ with continuous time $0\sim 1$), 
Eq. \ref{12} allows $s$ to take an arbitrary value in $[0,1]$. This is attributed to the analytic form of $\mathbf{h}_{t}$, which describes the proposed image-to-zero process.  % DDM to use an arbitrary sampling step size $s$. 
Technically speaking, Eq.~\ref{12} even allows us to generate images in a single step by directly setting $s=t=1$. Nevertheless, it is challenging to give a good estimate of $\boldsymbol{\phi}$ at the starting time. Moreover, one-step generation ($s=t$) causes the variance $\widetilde{\sigma}^{2}$ in Eq. \ref{12} to be zero. The two problems make the generated images appear blurry and lack diversity. Therefore, we still sample iteratively but use a much larger step size than DPMs: this drastically reduces the sampling steps from 1000 to 10 while maintaining relatively good image quality. Specifically, we commence at $t=1$ and take uniform steps of size $s$ until we reach $t=0$. The algorithmic details are outlined in Alg.~\ref{alg:2}. Detailed proof of Eq.~\ref{12} is provided in Section I-B of Supplementary Material.%Appendix~\ref{supsec:1.3}.%Section 1.2 of Supplementary Material.

% \begin{algorithm}[htb]  
%   \caption{Sampling algorithm of DDM.}  
%   %\label{alg:Framwork}  
%   \begin{algorithmic}[1]  
%     % \Require  
%     %   The proportion of the operation $o(\cdot)$ in connection from node $j$ to node $i$, $\gamma^{o}_{i,j}$, $j\leq i$;
%     % \Ensure  
%     %   The remaining operations list, $remain_o$;  
%     \State \textbf{Initialize} $\mathbf{x}_{t}\sim \mathcal{N}(\mathbf{0}, \mathbf{I}),t=1, N=num\_steps, s=1/N, \mathrm{Net}_{\boldsymbol{\theta}}$;  
%     %\label{code:fram:extract}  
%     \While {$t>0$}
%         \State $\boldsymbol{\phi}_{\boldsymbol{\theta}}, \boldsymbol{\epsilon}_{\boldsymbol{\theta}} = \mathrm{Net}_{\boldsymbol{\theta}}(\mathbf{x}_{t}, t)$;
%         \State $\mathbf{H}_{t} = \int_{0}^{t} {\mathbf{h}^{\boldsymbol{\phi}}_{t} \mathrm{d}t}$;
%         \State $\mathbf{\widetilde{u}} = \mathbf{x}_{t} + \mathbf{H}_{t-s}-\mathbf{H}_{t}-\frac{s}{\sqrt{t}}\boldsymbol{\epsilon}_{\boldsymbol{\theta}}$;
%         \State $\widetilde{\sigma} = \sqrt{\frac{s(t-s)}{t}}$;
%         \State $\mathbf{x}_{t} = \mathbf{\widetilde{u}} + \widetilde{\sigma}*\boldsymbol{\widetilde{\epsilon}}$,\ \ $\boldsymbol{\widetilde{\epsilon}}\sim \mathcal{N}(\mathbf{0}, \mathbf{I})$;
%         \State $t = t-s$;
%     \EndWhile; \\
%     \Return $\mathbf{x}_{t}$;  
%   \end{algorithmic}  
%   \label{alg:2}
% \end{algorithm}%\vspace{-15pt}

\begin{table*}[ht]
  \centering
%\small
\caption{Unconditional generative performances (FID$\downarrow$) on CIFAR-10 and CelebA-HQ-256 compared to previous DPMs including DDPM, SDE, LSGM, EDM, and CLD. Model size \emph{w.r.t} the number of parameters is shown.           \begin{tikzpicture}
    \fill[green] (0,0) rectangle (0.2,0.2);
    \end{tikzpicture} means FID is lower than the second-best method with statistical significance (p-value $<$ 0.05) based on the two sample t-test. \begin{tikzpicture}
    \fill[yellow] (0,0) rectangle (0.2,0.2);
    \end{tikzpicture} means the difference between our method and the closest best method is not statistically significant. For a fair comparison, we use the Euler-Maruyam solver to solve the reversed process of all diffusion models except for DDIM \cite{(3)pmlr-v139-nichol21a}, an ordinary differential equation-based accelerator. Note that we keep the model size of DDIM similar to ADM for a fair comparison.
    %denotes\textbackslash{}comparable with the second-high method with statistical significance (p-value $<$ 0.05) based on the two sample t-test. 
    % We did not run the 2000-step sampling on CelebA-HQ-256, because it takes more than 7 days on an RTX 3090 GPU.
    }%\vspace{-10pt}  
    \label{tab:1}
\resizebox{0.95\textwidth}{!}{
  \begin{tabular}{lcccccccc}
\hline
Dataset                                       & Method\textbackslash{}NFE & Param & 10     & 20     & 30  &  40  & 50 & 100 \\ \hline
\multicolumn{1}{l}{\multirow{7}{*}{CIFAR-10}} & DDPM \cite{(2)ddpm}       & 36M  & 296.84 & 140.68 & 101.46 & 48.53 & 38.36 & 19.87    \\
\multicolumn{1}{c}{}                          & SDE \cite{(9)sde} & 103M  & 304.73 & 193.94 & 141.41 & 90.82 & 66.32 & 28.26 \\
% \multicolumn{1}{c}{}                          & DDIM \cite{(3)pmlr-v139-nichol21a} & 36M  & 13.36 & 6.84 & 5.79 & 5.05 & 4.67 & 4.16 \\
\multicolumn{1}{c}{}                          & DDIM \cite{(3)pmlr-v139-nichol21a} & 88M  & 12.45 & 6.63 & 5.43 & 5.01 & 4.66 & 4.11 \\
\multicolumn{1}{c}{}                          & LSGM \cite{(27)DBLP:conf/nips/VahdatKK21} & 470M  & 27.64  & 11.69  & 7.33 & 5.80 & 5.19 & 3.99    \\
\multicolumn{1}{c}{}                          & CLD \cite{(25)DBLP:conf/iclr/DockhornVK22} & 103M  & 415.36 & 162.44 & 110.16 & 80.69 & 52.70 & 20.42 \\
\multicolumn{1}{c}{}                          & EDM \cite{(4)karras2022elucidating} & 54M  & 20.16 & 6.76 & 4.38 & 3.42 & 3.22 & 2.35 \\
\multicolumn{1}{c}{}                          & ADM (ours)                & 90M  & \textbf{10.74}\adjustbox{valign=c}{\begin{tikzpicture}[baseline=(current bounding box.center)]
\fill[green] (0,0) rectangle (0.2,0.2);
\end{tikzpicture}}  & \textbf{6.09}\adjustbox{valign=c}{\begin{tikzpicture}[baseline=(current bounding box.center)]
\fill[green] (0,0) rectangle (0.2,0.2);
\end{tikzpicture}}  & 4.33\adjustbox{valign=c}{\begin{tikzpicture}[baseline=(current bounding box.center)]
\fill[yellow] (0,0) rectangle (0.2,0.2);
\end{tikzpicture}}
& 3.41\adjustbox{valign=c}{\begin{tikzpicture}[baseline=(current bounding box.center)]
\fill[yellow] (0,0) rectangle (0.2,0.2);
\end{tikzpicture}}    & 3.22 & 2.81 \\ \hline
\multirow{5}{*}{CelebA-HQ-256}                & SDE \cite{(9)sde} & 62M     & 430.84 & 210.54 & 153.03 & 101.25 & 78.98 & 50.33 \\
                                              & LSGM \cite{(27)DBLP:conf/nips/VahdatKK21} & 470M     & 27.53  & 21.66 & 18.50 & 17.74 & 16.01 & 13.85 \\
                                              & LDM \cite{(23)DBLP:conf/cvpr/RombachBLEO22} & 305M     & 61.34  & 36.18 & 27.90 & 23.24 & 20.97 & 15.71 \\
                                              & CLD \cite{(25)DBLP:conf/iclr/DockhornVK22} & 103M     & 355.15 & 179.85 & 127.83  & 86.99 & 64.77  & 47.48 \\
                                              & EDM \cite{(4)karras2022elucidating} & 54M     & 30.29 & 19.26 & 13.33  & 10.88 & 9.34 & 8.02 \\
                                              & ADM (ours)                & 130M     & \textbf{26.10}\adjustbox{valign=c}{\begin{tikzpicture}[baseline=(current bounding box.center)]
\fill[green] (0,0) rectangle (0.2,0.2);
\end{tikzpicture}}   & \textbf{11.82}\adjustbox{valign=c}{\begin{tikzpicture}[baseline=(current bounding box.center)]
\fill[green] (0,0) rectangle (0.2,0.2);
\end{tikzpicture}}   & \textbf{8.72}\adjustbox{valign=c}{\begin{tikzpicture}[baseline=(current bounding box.center)]
\fill[green] (0,0) rectangle (0.2,0.2);
\end{tikzpicture}}   & \textbf{7.77}\adjustbox{valign=c}{\begin{tikzpicture}[baseline=(current bounding box.center)]
\fill[green] (0,0) rectangle (0.2,0.2);
\end{tikzpicture}} & \textbf{7.29}\adjustbox{valign=c}{\begin{tikzpicture}[baseline=(current bounding box.center)]
\fill[green] (0,0) rectangle (0.2,0.2);
\end{tikzpicture}} & \textbf{6.56}\adjustbox{valign=c}{\begin{tikzpicture}[baseline=(current bounding box.center)]
\fill[green] (0,0) rectangle (0.2,0.2);
\end{tikzpicture}} \\ \hline
\end{tabular} 
}
\end{table*}

\begin{figure*}[!ht]
    \centering
    %\fbox{\rule[-.5cm]{0cm}{4cm} \rule[-.5cm]{4cm}{0cm}}
    %\includegraphics[width=1\linewidth]{figures/framework.pdf}
    \begin{overpic}
    [width=1\textwidth]{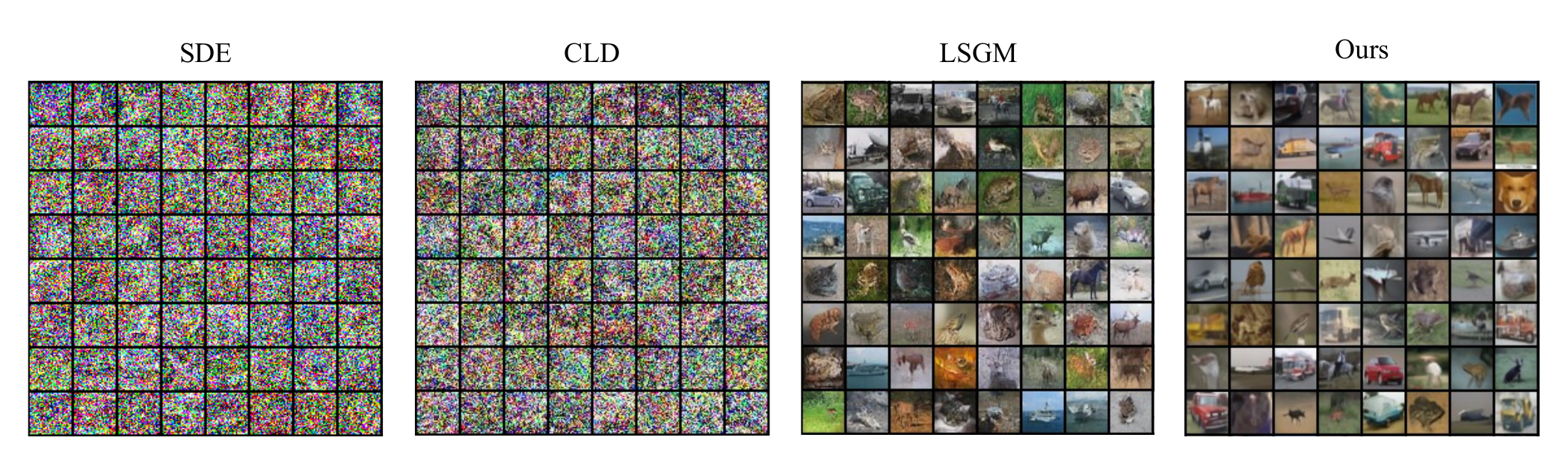}
    % \put(0.1, 21){\small Input}
    % \put(-0.5, 12){\small DDPM}
    % \put(0.2, 4.2){\small Ours}
    % \put(21.0, 0){\small JPPNet\cite{(25)liang2018look}}
    % \put(41.5, 0){\small MuLA\cite{(26)nie2018mutual}}
    \end{overpic} %\vspace{-15pt}
    \caption{Comparisons of 10-step unconditional generation on CIFAR10.}\label{supfig:2}\vspace{-5pt}
\end{figure*}

\subsection{Remarks}
% \textbf{Why decoupling the image-to-noise process into image to zero and zero to noise eases training?}
\textbf{Why simultaneous image-to-zero and zero-to-noise eases training?}
In the image-to-noise process, estimating the noise component from $\mathbf{x}_{t}$, which is deeply mixed by $\mathbf{x}_{0}$ and noise, is not an easy task. Compared to image-to-noise process that only uses the noise loss to supervise the network, image-to-zero and zero-to-noise mappings add the $\mathbf{x}_{0}$ modeling branch and use the noise and $\mathbf{x}_{0}$ to supervise the network simultaneously, thus reducing learning difficulty.

\textbf{Why ADM allows for fewer function evaluations?}
Previous DPMs directly model the image-to-noise process, which requires numerical integration with a very small step size since it is not analytic. Differently, ADM has an analytic image-to-zero process, allowing the reversed process to be solved with arbitrary step sizes, so we can generate samples with fewer function evaluations.

\textbf{Establishing a new diffusion framework vs. fast sampling.}
Our method is not directly comparable but complementary with fast sampling methods. ADMs benefit from the proposed analytic image-to-zero process to obtain few-step generative ability. Differently, existing fast sampling methods do not change the forward diffusion process but adopt high-order ODE solvers or distillation techniques for few-step generations. In fact, the latter can potentially be applied to ADMs.

\begin{table*}[ht] %We compare the proposed ADM with DDPM and other state-of-the-art methods.}
    \centering
    % \footnotesize
    \caption{Comparing ADM with DDPM and the state of the art on conditioned generation.}%\vspace{-10pt}
    \label{tab:2}
  \setlength{\tabcolsep}{3pt}

\resizebox{1\textwidth}{!}{
    \begin{tabular}{llllllll}
    \hline
    \multicolumn{2}{c}{Saliency Detection}                                                         & \multicolumn{2}{c}{Super-resolution}                                                   & \multicolumn{2}{c}{Image Inpainting}                                                       & \multicolumn{2}{c}{Edge Detection} \\ \hline
    Method                                                       & MAE$\downarrow$                 & Method                                               & PSNR$\uparrow$                  & Method                                                    & FID$\downarrow$                & Method             & ODS$\uparrow$ \\
    EVP \cite{(46)liu2023explicit}              & 0.026                           & SPSR \cite{(49)ma2020structure}     & 26.71                           & Co-mod GAN \cite{(53)zhao2021large}      & 19.4                           & PiDiNet \cite{(65)su2021pixel} & 0.789         \\
    Selfreformer \cite{(47)yun2022selfreformer} & 0.026                           & IDM \cite{(50)gao2023implicit}      & 27.10                           & AOT-GAN \cite{(54)zeng2022aggregated}    & 7.28                           & FCL-Net \cite{(66)xuan2022fcl} & 0.807         \\
    Tracer-3 \cite{(48)lee2022tracer}           & 0.028                           & FxSR \cite{(51)park2022flexible}    & 27.51                           & LaMa \cite{(55)suvorov2022resolution}    & 6.13                           & EDTER \cite{(67)pu2022edter} & 0.824         \\
    Tracer-7 \cite{(48)lee2022tracer}           & \textbf{0.022} & SROOE \cite{(52)park2023perception} & 27.69                           & Wavepaint \cite{(56)jeevan2023wavepaint} & 5.59                           & UAED \cite{(68)zhou2023treasure} & 0.829         \\ \hline
    DDPM-5 step                                                  & 0.497                           & DDPM-5 step                                          & 9.13                            & DDPM-5 step                                               & 114                            & DDPM-5 step        & 0.559         \\
    DDPM-10 step                                                 & 0.473                           & DDPM-10 step                                         & 12.46                           & DDPM-10 step                                              & 95.8                           & DDPM-10 step       & 0.563         \\
    ADM-5 step (Ours)                                            & 0.026                           & ADM-5 step (Ours)                                    & 27.61                           & ADM-5 step (Ours)                                         & 5.53                           & ADM-5 step (Ours)  & \textbf{0.834}         \\
    ADM-10 step (Ours)                                           & 0.025                           & ADM-10 step (Ours)                                   & \textbf{28.30} & ADM-10 step (Ours)                                        & \textbf{5.25} & ADM-10 step (Ours) & \textbf{0.834}         \\ \hline
    \end{tabular}
}
\end{table*}

\section{Experiments}\label{sec:exp}
    \subsection{Experimental Setup}\label{sec:exp.1}
    \textbf{Implementation Details.}
    We perform experiments on both unconditioned and conditioned image generation. % and conduct conditional image generation tasks to demonstrate the generalizability of our approach in various scenarios. 
    For unconditional image generation, we evaluate ADM on the low-resolution CIFAR-10 dataset \cite{(19)krizhevsky2009learning} as well as the high-resolution dataset CelebA-HQ-256 \cite{(20)DBLP:conf/iclr/KarrasALL18}. For conditional image generation, we consider four tasks: image inpainting on CelebA-HQ-256, image super-resolution on DIV2K \cite{(21)DBLP:conf/cvpr/AgustssonT17}, saliency detection on DUTS \cite{(45)wang2017learning} and edge detection on BSDS \cite{(64)arbelaez2010contour}. Due to the limitations of computing resources, we translate the diffusion process into latent space for high-resolution ($>$256) image synthesis following \cite{(23)DBLP:conf/cvpr/RombachBLEO22}. %We follow \cite{(9)sde} using NCSNv2 \cite{(24)song2020improved} as the basic model. 
    For the architecture, since ADM aims to predict the image and noise components separately, we modify the original U-Net architecture by adding an extra decoder as shown in Fig.~\ref{supfig:1}. For conditioned image generation, we use an additional encoder to extract features for the conditional input which are used to correlate the features of $\mathbf{x}_{t}$ by attention mechanisms. For the selection of $\mathbf{h}_{t}$, we use a simple yet effective function $\mathbf{h}_{t}=\mathbf{c}$.%, where $\mathbf{c}$ is a learnable parameter. % and more discussion can be found in Sec.~\ref{sec:ablation}.

\textbf{Architecture of Network.} As shown in Fig.~\ref{supfig:1}, we modify the original U-Net architecture \cite{(24)song2020improved} and add an extra decoder so that our model has two outputs for predicting image and noise components respectively. For conditioned generation tasks, we utilize a down-sampling encoder like the U-Net encoder to extract multi-level features of the conditioned input, and concatenate these features and the image features with the same levels as the decoder's inputs. In practice, we use the Swin-B \cite{(57)liu2021swin} as our condition encoder.

%\textbf{Training.} We use the AdamW optimizer with scheduled learning rates and train ADM for 800k iterations (training details can be found in Section 2.3 of Supplementary Material).
\textbf{Training Hyper-parameters.} We use the AdamW optimizer with scheduled learning rates to train our model. We provide an overview of the hyper-parameters of all trained ADMs in Tab.~\ref{subtab:1}. Different from previous models that usually adopt a constant learning rate, we implement the polynomial policy to decay the learning rate gradually, which can be formulated by:
$\gamma=\max(\gamma_{0}\cdot (1-N_{iter}/N{total})^{p}, \gamma_{min})$. Here $\gamma_{0}$ is the initial learning rate and $\gamma_{min}$ denotes the smallest learning rate, $N_{iter}$ and $N_{total}$ correspond to the current iteration number and total iteration number, $p$ a hyper-parameter and we set it to 0.96. Additionally, we employ the exponential moving average (EMA) to prevent unstable model performances during the training process. We have observed that using mixed-precision (FP16) training negatively impacts the generative performances, hence, we do not utilize it.

\textbf{Obtaining Ground Truth of $\boldsymbol{\phi}_{\boldsymbol{\theta}}$}. In the training phase, we obtain the ground truth by solving $\mathbf{x}_{0} + \int_{0}^{1} {\mathbf{h}_{t}\mathrm{d}t}=\mathbf{0}$. For a simple example $\mathbf{h}_{t}=\mathbf{c}$, the only parameter of $\mathbf{h}_{t}$ is $\mathbf{c}$ and we can easily get: $\mathbf{c} = -\mathbf{x}_{0}$. Thus, the ground truth of $\boldsymbol{\phi_{\boldsymbol{\theta}}}$ is $-\mathbf{x}_{0}$. For the linear function $\mathbf{h}_{t}=\mathbf{a}t+\mathbf{c}$, we can not solve the two parameters $\mathbf{a}, \mathbf{c}$ using one equation. To avoid this problem, we first sample one of parameters from $\mathcal{N}(\mathbf{0}, \mathbf{I})$, and substitute it into $\mathbf{x}_{0} + \int_{0}^{1} {\mathbf{h}_{t}\mathrm{d}t}=\mathbf{0}$ to solve another parameter. In this way, we concatenate $\mathbf{a}, \mathbf{b}$ as the ground truth of $\boldsymbol{\phi_{\boldsymbol{\theta}}}$. The ground truths of other functions can be solved in a similar way.

\textbf{Final Denoising Step.} In general, the generated samples typically contain small noise that is hard to detect by humans \cite{(44)DBLP:conf/iclr/Jolicoeur-Martineau21}. To remove this noise, we follow \cite{(9)sde} letting the last denoising step occur at $t=\Delta t$ where $\Delta t$ is the smallest step size.

\textbf{Evaluation.} We use the Fréchet inception distance (FID) to measure image sample quality for unconditioned image generation. We generate 50,000 and 30,000 samples for CIFAR-10 and Celeb-A-HQ respectively to calculate FID. For super-resolution, we use the mean peak signal to noise ratio (PSNR) of the test set for evaluation. For image inpainting, we follow \cite{(55)suvorov2022resolution} to use 26,000 images for training and 2,000 images for testing, and report FID metric. For saliency detection, we calculate the mean absolute error (MAE) on the whole test set. For edge detection, we report the F-scores of Optimal Dataset Scale (ODS) to evaluate the performance. We also record the number of function evaluations (NFEs) during synthesis when comparing with previous DPMs.%\vspace{-10pt}

\begin{figure*}
  \centering
  %\fbox{\rule[-.5cm]{0cm}{4cm} \rule[-.5cm]{4cm}{0cm}}
  %\includegraphics[width=1\linewidth]{figures/framework.pdf}
  %\vspace{-10pt}
  \begin{overpic}
    [width=1\textwidth]{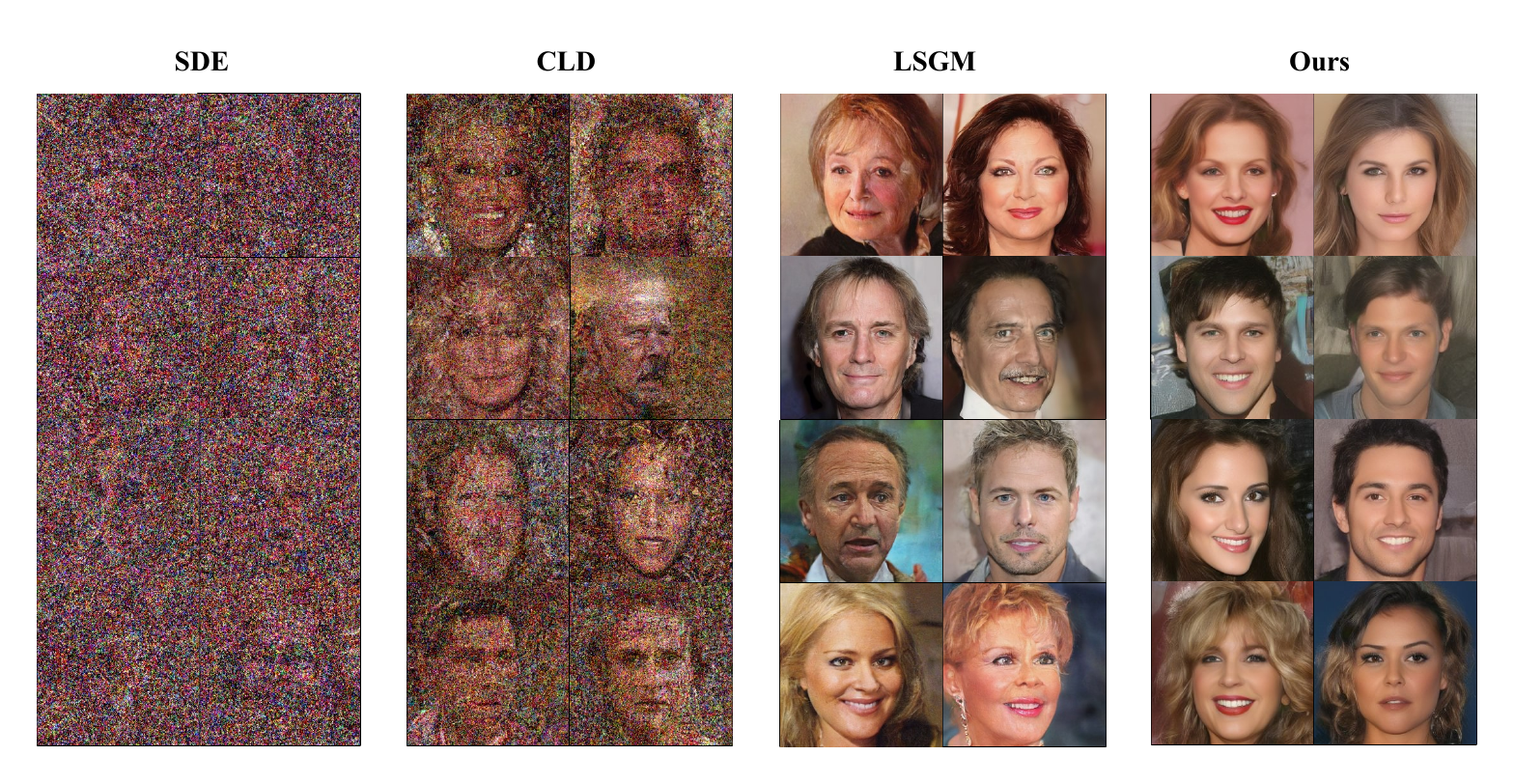}
    % \put(83, 31.5){\small $t$}
    % \put(83, 11.5){\small $t$}
    % \put(69.6, 35.6){\small $\boldsymbol{h}_t$}
    % \put(21.0, 0){\small JPPNet\cite{(25)liang2018look}}
    % \put(41.5, 0){\small MuLA\cite{(26)nie2018mutual}}
  \end{overpic}%\vspace{-10pt}
  \caption{Qualitative method comparison of unconditioned image generation with 10 function evaluations on CelebA-HQ-256. SDE, CLD, LSGM, and the proposed ADM are compared.}\label{fig:3} %\vspace{-10pt}
\end{figure*} 

\begin{figure}
    \centering
    \includegraphics[width=\linewidth]{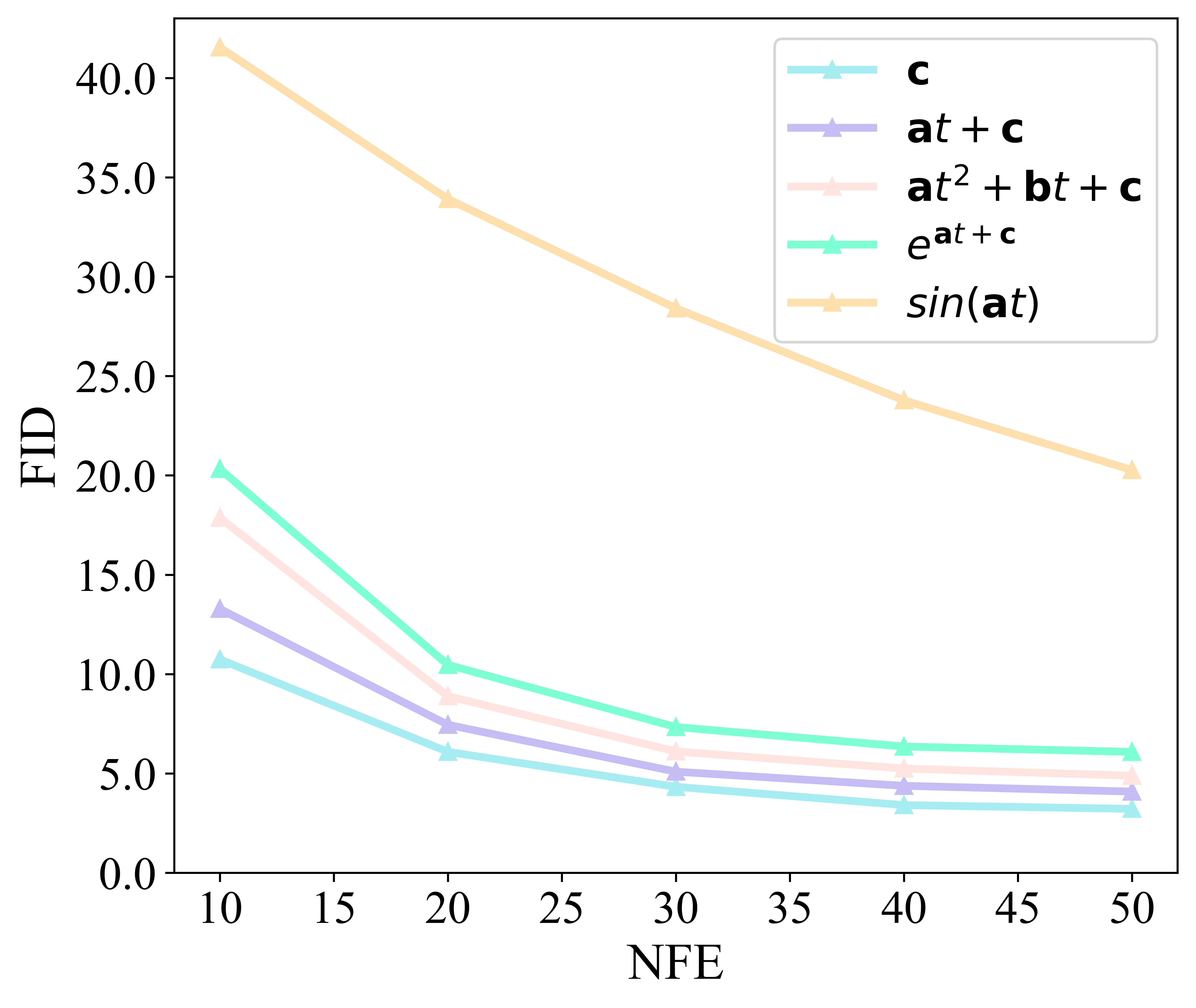}
    \caption{Comparing different choices of $\mathbf{h}_{t}$ on CIFAR-10. FID for unconditioned image generation is used as a metric. We report FID under 10$\sim$50 steps.}
    \label{fig:ht}
\end{figure}

\begin{figure*}[h]
    \centering
    \includegraphics[width=1\textwidth]{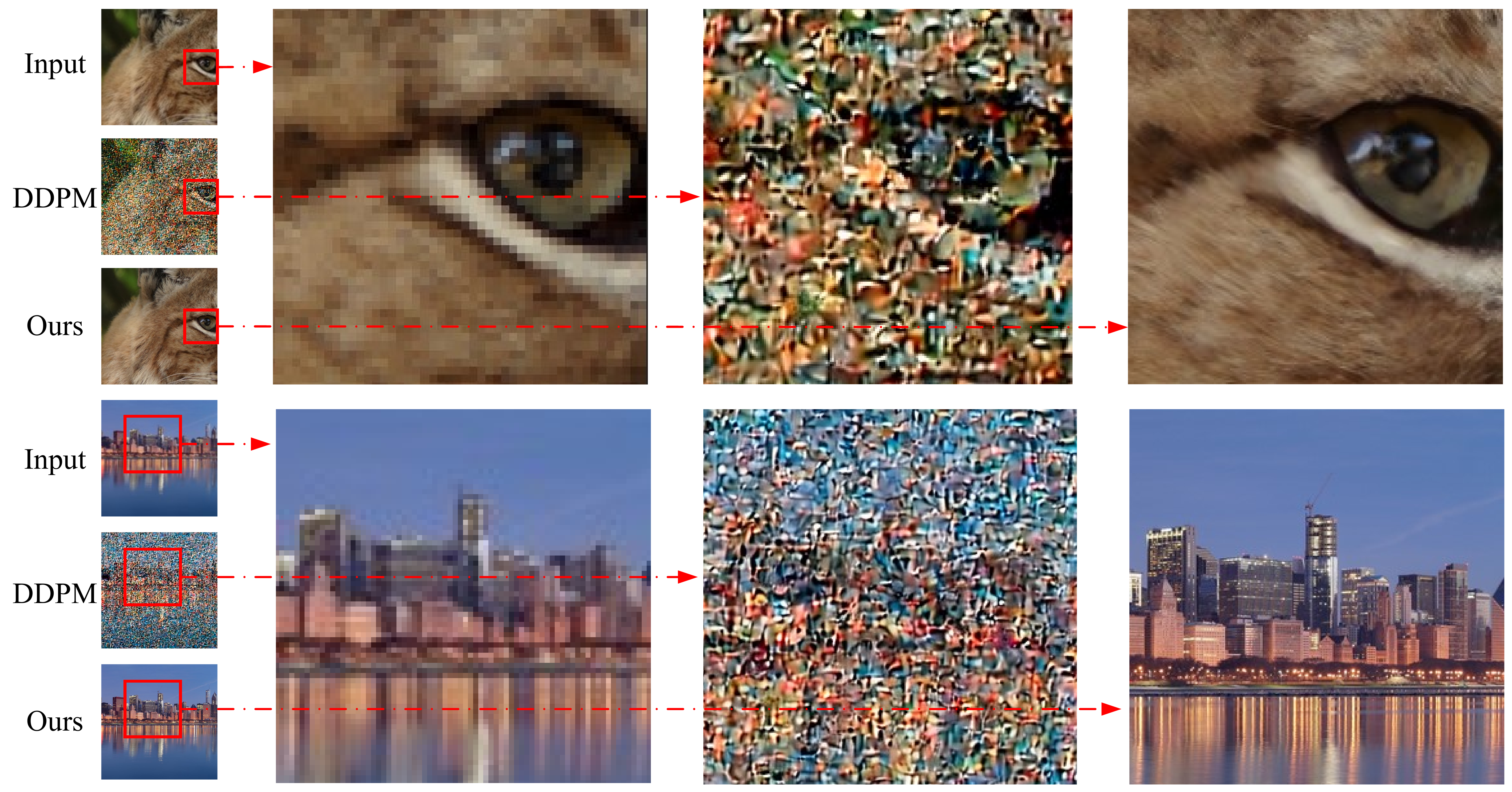}
    \caption{10-step super-resolution visulization.}%\vspace{-10pt}
    \label{fig:5}\vspace{0pt}
\end{figure*}

\begin{table*}[t]
\caption{Comparing ADM and DDPM on conditioned generation with various steps. We report MAE $\downarrow$, PSNR $\uparrow$, and FID $\downarrow$ for saliency detection, super-resolution, and image inpainting, \textit{resp.}}
    \label{tab:3}
    %\footnotesize
    \setlength{\tabcolsep}{3pt}
    \centering
    \resizebox{\textwidth}{!}{
    \begin{tabular}{l|cccc|cccc|cccc|llll}
\hline
                                           & \multicolumn{4}{c|}{Saliency detection} & \multicolumn{4}{c|}{Super-resolution} & \multicolumn{4}{c|}{Image inpainting} & \multicolumn{4}{c}{Edge Detection}                                                                 \\ \hline
Method\textbackslash{}NFE & 10       & 20       & 50      & 100     & 10      & 20      & 50      & 100     & 10      & 20      & 50      & 100     & \multicolumn{1}{c}{10} & \multicolumn{1}{c}{20} & \multicolumn{1}{c}{50} & \multicolumn{1}{c}{100} \\
DDPM                                       & 0.4731   & 0.3846   & 0.3059  & 0.1985  & 12.46   & 13.58   & 14.29   & 15.91   & 95.8    & 89.9    & 71.6    & 64.3    & 0.563                  & 0.567                  & 0.582                  & 0.763                   \\
ADM                                        & 0.0253   & 0.0251   & 0.0252  & 0.0250  & 28.30   & 28.29   & 28.35   & 28.33   & 5.25    & 5.19    & 5.15    & 5.15    & 0.834                  & 0.833                  & 0.834                  & 0.835                   \\ \hline
\end{tabular}
}
\end{table*}

\subsection{Unconditioned Generation}
We present results and compare methods in Tab.~\ref{tab:1}. Two observations are made below. 

    % \begin{table}
    %   \caption{Unconditional CelebA-HQ-256 generative performances (FID$\downarrow$) compared to previous DPMs.} 
    %   \label{tab:2}
    %   \centering
    %   \begin{tabular}{lccccc}
    %     \hline
    %     Method \textbackslash NFE & 10     & 20     & 50     & 1000 \\ \hline
    %     %DDPM                      & 296.84 & 140.68 & 38.36  & 3.17   \\ 
    %     SDE                       & 430.84 &  &  & 7.23\\ 
    %     LSGM                      & 27.53  & 14.39 &  & -\\ 
    %     CLD                       & 355.15 &        &    & -\\ 
    %     ADM (Ours)                & \textbf{27.45}  & \textbf{14.37}   & \textbf{10.52}   & \textbf{7.10} \\ \hline
    %   \end{tabular}
    % \end{table}

\textbf{ADM is very competitive in long-step image generation.} 
On the low-resolution CIFAR-10 dataset, ADM gets very competitive performances with the state-of-the-art method on 50-step and 100-step settings. On the high-resolution CelebA-HQ-256 dataset, ADM achieves FID of 7.29 and 6.56, respectively, using 50 and 100 function evaluations. Particularly, compared to the baseline DDPM \cite{(2)ddpm}, ADM requires only $1/20$ of the function evaluations to generate images of comparable quality. The results demonstrate the validness and the great ability of the ADM framework for synthesizing high-fidelity images.
% On the CIFAR-10 and  datasets, ADM achieves FID of 2.19 and 7.10, respectively, using 2,000 and 1,000 function evaluations. This is very competitive compared with SDE, SLGM, CLD, and DDPM and demonstrates the validness of the ADM framework. Note that LSGM uses a much more lager model with 470M parameters while ADM only has 175M. 

\textbf{ADM has superior performance in few-step generation.} On the CIFAR-10 dataset, ADM consistently outperforms existing methods under 10 and 20 steps. Using 10 steps, FID of our method is 10.74 much lower than the second best method EDM. On CelebA-HQ-256, our method also outperforms all comparable methods under all the few-step scenarios. Note that for fair comparisons, we use Euler-Maruyama \cite{(26)DBLP:conf/para/ArtemievGM95} to solve the competing diffusion models. More qualitative results are provided in Fig.~\ref{supfig:2} and Fig.~\ref{fig:3}, where our method is significantly better than SDE and CLD under 10 function evaluations and on par with LSGM. %The comparison with DDPM in super-resolution is shown in Fig.~\ref{fig:5}.
It is worth noting that our method outperforms the ordinary differential equation-based accelerator DDIM \cite{(3)pmlr-v139-nichol21a} using the vanilla Euler-Maruyam solver. % which only optimizes the sampling process but training. Differently, our method directly optimize the training 

\subsection{Conditioned Generation}

\begin{table}[t]
    \centering
\setlength{\tabcolsep}{7pt}
    \centering
    \caption{Hyperparameter, architecture, and loss analysis. $t$ is defined in Eq. \ref{7}. FID for 10-step unconditioned generation on CIFAR-10 is reported.%, U-Net architecture and loss type.
    }\label{tab:5}%\vspace{-15pt}
    \resizebox{\linewidth}{!}{
    \begin{tabular}{lccc}
\hline
  &  $\lambda_{1}$               & $\lambda_{2}$          & 10-step FID          \\\hline
\multirow{4}{*}{\begin{tabular}[c]{@{}l@{}}Loss weight\end{tabular}} & 1                    & 1                    & 12.19                \\
  &1/$t$          & 1/($1-t$)            & 11.07                \\
  &$e^{t}$               & $e^{1-t}$          & 11.56                \\ 
  &$\frac{t^{2}-t+1}{t}$               & $\frac{t^{2}-t+1}{(1-t)^{2}}$          & 10.74             \\\hline
\multirow{2}{*}{Architecure}                                              & \multicolumn{2}{c}{U-Net+conv}              & 12.63                \\
& \multicolumn{2}{c}{U-Net+decoder}           & 10.74                \\ \hline
\multirow{3}{*}{\begin{tabular}[c]{@{}l@{}}Loss type\end{tabular}}      & \multicolumn{2}{c}{$\ell_1$}                      & 12.08                \\
& \multicolumn{2}{c}{$\ell_2$}                      & 10.74                \\
& \multicolumn{2}{c}{$\ell_1+\ell_2$}                      & 11.57                \\\hline
& \multicolumn{1}{l}{} & \multicolumn{1}{l}{} & \multicolumn{1}{l}{}
\end{tabular}%\vspace{-10pt}
    }
\end{table}

\begin{table}[t]
    \centering
    \caption{Ablations on components of ADM. `AIA' means `analytical image attenuation'. Note that we reproduce DDPM with a larger network parameter.}
    \label{tab:6}
    \resizebox{0.9\linewidth}{!}{
    \begin{tabular}{lccc}
    \hline
Method\textbackslash{}Step & 10 & 20 & 30 \\\hline
DDPM                       & 274.47  & 130.55   &  89.25  \\
with $\mathbf{x}_{0}$ prediction   &  203.51  & 100.09 & 70.81 \\
with AIA  &  13.17   & 8.85   & 5.37 \\
with both  &  10.74   & 6.09   & 4.33 \\
\hline 
\end{tabular}
}
\end{table}

\begin{figure}
    \centering
    %\fbox{\rule[-.5cm]{0cm}{4cm} \rule[-.5cm]{4cm}{0cm}}
    %\includegraphics[width=1\linewidth]{figures/framework.pdf}
    \begin{overpic}
    [width=1\linewidth]{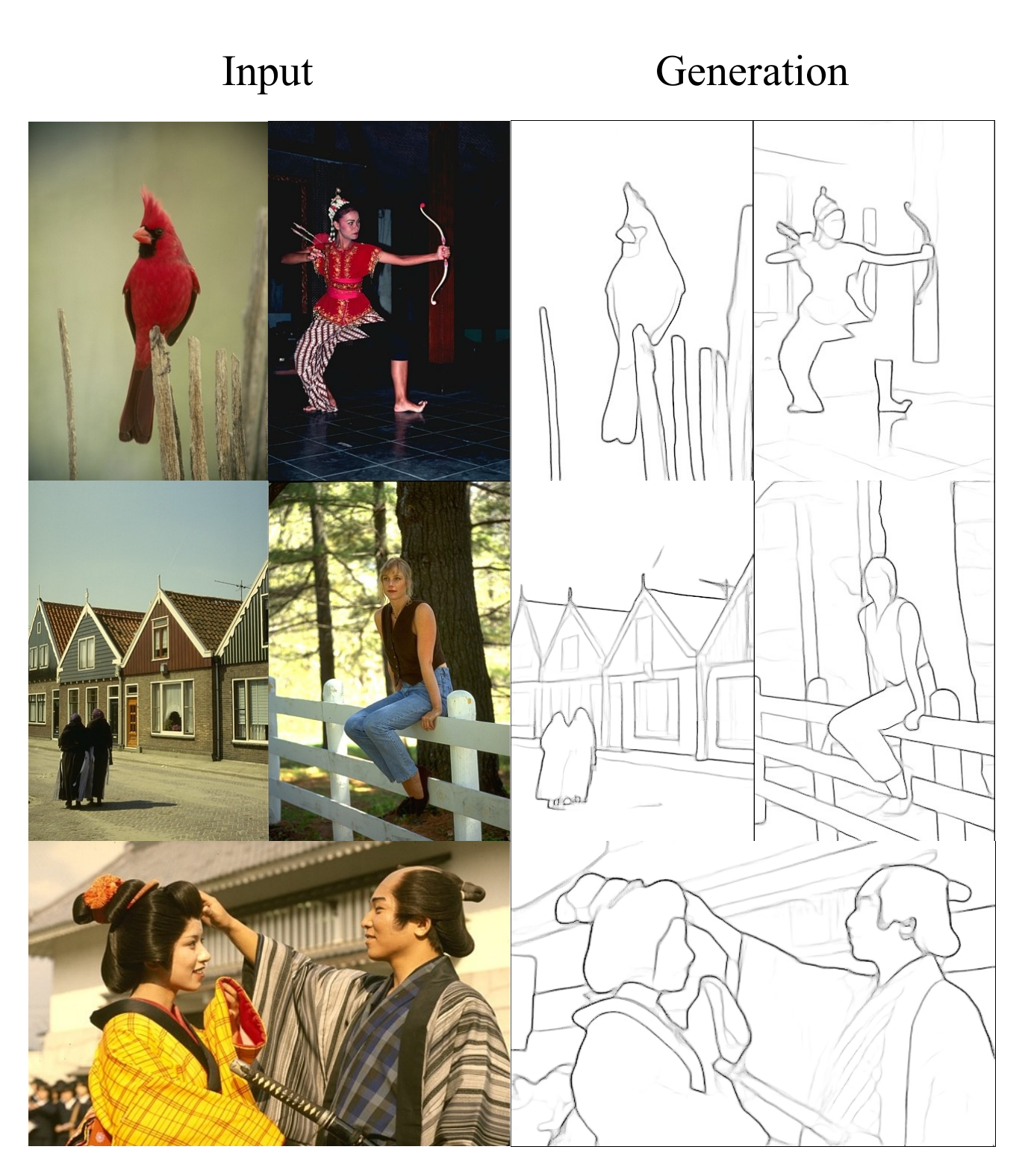}
    % \put(0.1, 21){\small Input}
    % \put(-0.5, 12){\small DDPM}
    % \put(0.2, 4.2){\small Ours}
    % \put(21.0, 0){\small JPPNet\cite{(25)liang2018look}}
    % \put(41.5, 0){\small MuLA\cite{(26)nie2018mutual}}
    \end{overpic} %\vspace{-15pt}
    \caption{Quantitative results of edge detection.}\label{fig:edge}\vspace{-5pt}
\end{figure}

\begin{figure}
    \centering
    %\fbox{\rule[-.5cm]{0cm}{4cm} \rule[-.5cm]{4cm}{0cm}}
    %\includegraphics[width=1\linewidth]{figures/framework.pdf}
    \begin{overpic}
    [width=1\linewidth]{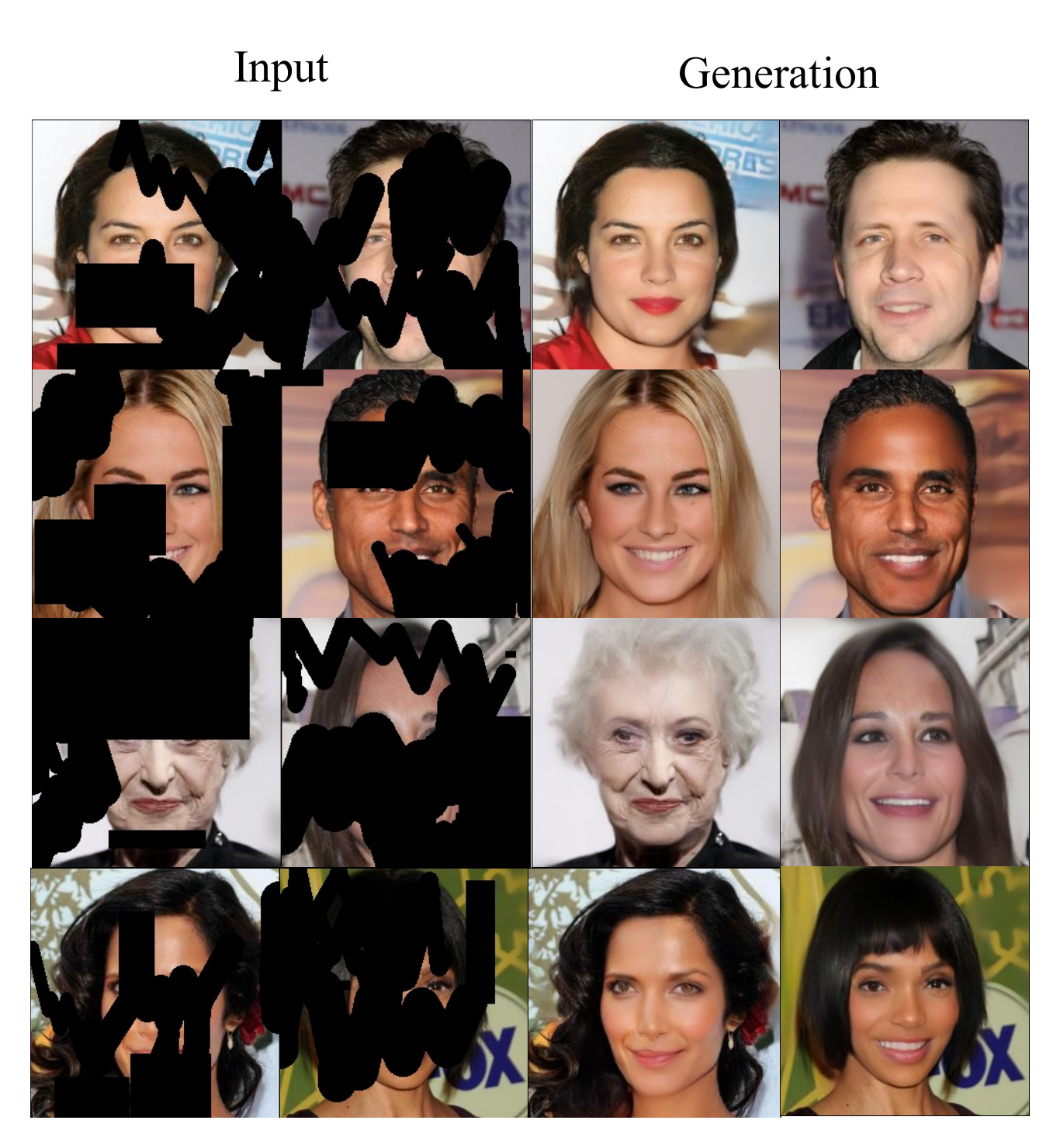}
    % \put(0.1, 21){\small Input}
    % \put(-0.5, 12){\small DDPM}
    % \put(0.2, 4.2){\small Ours}
    % \put(21.0, 0){\small JPPNet\cite{(25)liang2018look}}
    % \put(41.5, 0){\small MuLA\cite{(26)nie2018mutual}}
    \end{overpic} %\vspace{-15pt}
    \caption{Quantitative results of image inpainting.}\label{fig:inpaint}\vspace{-5pt}
\end{figure}

\begin{figure}
    \centering
    %\fbox{\rule[-.5cm]{0cm}{4cm} \rule[-.5cm]{4cm}{0cm}}
    %\includegraphics[width=1\linewidth]{figures/framework.pdf}
    \begin{overpic}
    [width=1\linewidth]{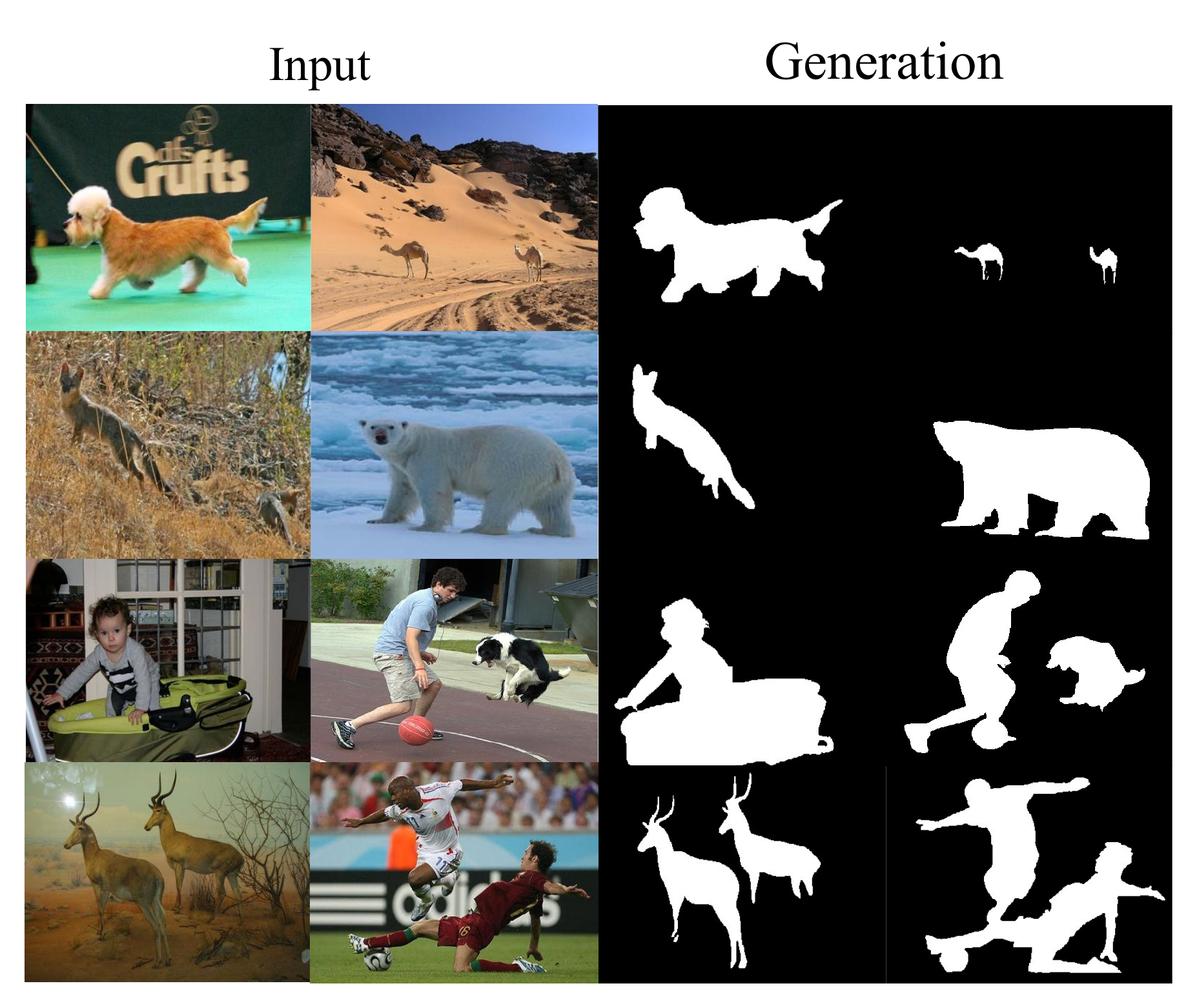}
    % \put(0.1, 21){\small Input}
    % \put(-0.5, 12){\small DDPM}
    % \put(0.2, 4.2){\small Ours}
    % \put(21.0, 0){\small JPPNet\cite{(25)liang2018look}}
    % \put(41.5, 0){\small MuLA\cite{(26)nie2018mutual}}
    \end{overpic} %\vspace{-15pt}
    \caption{Quantitative results of saliency detection.}\label{fig:saliency}\vspace{-5pt}
\end{figure}

\textbf{Conditioned image generation with 5 or 10 steps.} 
We compare the proposed ADM with both DDPM and state-of-the-art methods for conditioned image generation.
\begin{itemize}
\item \textbf{Saliency detection}. %aims to fill the masked regions of an image with new content matching the unmasked regions. 
As shown in Tab.~\ref{tab:2}, ADM shows very superior performances to DDPM for few-step generations. Moreover, ADM achieves very competitive performance with state-of-the-art methods within 10-step generation. Note that TRACER-7 gets the best result by using a much larger training size (640$\times$640) than other comparisons (384$\times$384).
\item \textbf{Super-resolution}. Compared to state-of-the-art methods, ADM only takes 5 NFEs to get very comparable performance, and achieves the best PSNR of 28.30 through 10 NFEs. The comparison with DDPM in super-resolution is shown in Fig.~\ref{fig:5}. 
\item \textbf{Image inpainting}. We can observe that ADM has great advantages in image inpainting that it only uses 5 NFEs to exceed other state-of-the-art methods. Compared to vanilla DDPM, ADM exceeds it significantly with 10-step sampling. 
\item \textbf{Edge detection}. The proposed ADM obtains the best ODS of 0.834, outperforming previous state-of-the-art methods with only 5 steps. According to the visual results in Fig.~\ref{fig:edge}, our method can generate accurate and crisp edge predictions.
\end{itemize}
The above experiments demonstrate the effectiveness of the proposed ADM and it can be easily applied to downstream image-conditioned tasks. More quantitative results of edge detection, image inpainting, and saliency detection can be found in Fig.~\ref{fig:edge}, Fig~\ref{fig:inpaint} and Fig~\ref{fig:saliency}.

\textbf{Conditioned image generation with more steps.}
As shown in Tab.~\ref{tab:3}, we evaluate the long-step performance on image-conditioned tasks. Different from unconditioned generation, the improvement of increasing NFE is very slight for ADM. We believe that the performances are easily saturated since the conditioned input provides strong prior information, letting ADM converge rapidly.

%\textbf{Semantic Map to Image.} Semantic map to image aims to convert the semantic map containing only the category information of each pixel into a real RGB image. We select the Cityscapes dataset which contains 20 classes of the urban scenes to generate 512$\times$1024 resolution images, which is much challenging. In Fig.~\ref{fig:6}, we showcase the results of our method and DDPM, with only 10 function evaluations. The generated images exhibit clear contours for different objects, demonstrating the effectiveness of our approach. In comparison, the results obtained by DDPM appear vague.
% \begin{figure}
%   \centering
%   %\fbox{\rule[-.5cm]{0cm}{4cm} \rule[-.5cm]{4cm}{0cm}}
%   %\includegraphics[width=1\linewidth]{figures/framework.pdf}
%   \begin{overpic}
%     [width=1.0\textwidth]{figures/cityscapes.pdf}
%     % \put(83, 31.5){\small $t$}
%     % \put(83, 11.5){\small $t$}
%     % \put(69.6, 35.6){\small $\boldsymbol{h}_t$}
%     % \put(21.0, 0){\small JPPNet\cite{(25)liang2018look}}
%     % \put(41.5, 0){\small MuLA\cite{(26)nie2018mutual}}
%   \end{overpic} \vspace{-10pt}
%   \caption{Qualitative results of 10-step generation on semantic map to image task.}\label{fig:6} %\vspace{-15pt}
% \end{figure} 

\subsection{Hyper-parameter Analysis}\label{sec:ablation} 

\textbf{Comparing different choices of $\mathbf{h}_{t}$.}
We conduct the analysis experiments on CIFAR-10 dataset. Eq.~\ref{7} shows that we can construct the forward process with arbitrary explicit functions which are obtained by solving $\mathbf{x}_{0} + \int_{0}^{1} {\mathbf{h}_{t}\mathrm{d}t}=\mathbf{0}$. Here, we compare five functions shown in %such as constant function, linear function, quadratic function and exponential function and compare their performances. As shown in 
Fig.~\ref{fig:ht}. The constant function has the lowest FID under the three NFEs. The reason might be that the constant function only has one learnable term $\mathbf{c}$, reducing the learning difficulty for the model. On the other hand, the constant function implies that the image component undergoes uniform attenuation during the forward process, providing a clearer and more concise learning pathway. The sine function limits the value range of $\mathbf{h}_{t}$ to $[-1, 1]$, which hinders the modeling of image-to-zero process.

\textbf{Hyperparameter sensitivity analysis.} Here we assess the weighting hyperparameters $\lambda_{1}$ and $\lambda_{2}$ defined in Section~\ref{sec:4.3}. We set the baseline weights to: $\lambda_{1} = \lambda_{2} = 1$. Since $\mathbf{x}_{t}$ is closer to $\boldsymbol{\epsilon}$ when $t\rightarrow 1$, the second term in Eq.~\ref{11} is much smaller than the first term when $t\rightarrow 1$. The opposite phenomenon can be observed for the first term when $t\rightarrow 0$. Therefore, we empirically set two types of weighting selection to balance the two terms. Besides, we follow \cite{(4)karras2022elucidating} to derive the adaptive weights:  $\lambda_{1} = \frac{t^{2}-t+1}{t}, \lambda_{2} = \frac{t^{2}-t+1}{(1-t)^{2}}$. As shown in Tab.~\ref{tab:5}, the time-dynamic weights can slightly improve the baseline.  %It would be interesting to study this in future.

\textbf{Comparing variants of the U-Net architecture.}
%The original U-Net architecture only has one decoder for one-output prediction. 
Because ADM predicts both the noise and $\mathbf{x}_0$, %there are two predictions in ADM, 
we need to modify the architecture so that the modified U-Net has two outputs. There are two variants: \textbf{(1)} we use an additional branch, which consists of two stacked 3x3 convolutional layers with stride 2 and padding 1, from the feature map before the output layer; \textbf{(2)} we add a new decoder branch that has the same architecture as the original decoder to obtain the another output. We named the two architectures U-Net+conv and U-Net+decoder. Tab.~\ref{tab:5} shows U-Net+decoder performs better.

\textbf{Comparing $\ell_1$ and $\ell_2$ loss in Eq.~\ref{11}.}
We also analyze the effect of $\ell_1$ and $\ell_2$ loss for supervising the network. Results in Tab.~\ref{tab:5} indicate that the $\ell_2$ loss is more suitable compared with the $\ell_1$ loss.

\textbf{Ablation on components of ADM}. As reported in Tab,~\ref{tab:6}, we ablate the effects of ADM's components on CIFAR-10, including $\mathbf{x}_{0}$ prediction branch and the analytical image attenuation. `with $\mathbf{x}_{0}$ prediction' means we use the image component to supervise the network, whereas `with analytical image attenuation' denotes we incorporate the analytical image attenuation process into the forward process but do not use the image component as supervision. With simultaneous image-to-zero and zero-to-noise, we can improve the few-step generative quality significantly since the analytical image attenuation allows the reversed-time SDE to be solved with an arbitrary step size, thus maintaining the generative quality with much fewer denoising steps. On the other hand, the $\mathbf{x}_{0}$ branch provides additional supervision for the network, which can further improve the generative performances.

\textbf{Computational time cost.} We use a computing server with Intel Xeon Silver 4210R CPU and RTX3090 GPU. The training time on CIFAR-10 and CelebA-HQ-256 is around 20 GPU days and 30 GPU days, respectively. The inference time to generate one batch of 16 images on CIFAR-10 and CelebA-HQ-256 is 1.05 second and 1.98 second in 10-step setting, respectively. We also calculate the Floating-point Operations (FLOPs) to compare the computational cost of different methods. For generating a 32$\times$32 image, our model takes 18G FLOPs, while DDPM, SDE, CLD, EDM and LSGM take 7G, 12G, 13G, 10G and 31G respectively, which indicates out model achieves the best FID metrics for 10-50 step sampling with a comparable computational cost.

% \begin{table}
%   \caption{Ablation studies of different explicit transition probabilities on CIFAR-10.} 
%   \label{tab:5}
%   \centering
%   \begin{tabular}{lccccc}
%     \hline
%     $\mathbf{h}_{t}$ \textbackslash NFE & 10     & 50     & 1000 \\ \hline
%     $\mathbf{c}$                     & \textbf{10.52}  & \textbf{6.19}    & \textbf{2.40} \\ 
%     $\mathbf{a}t+\mathbf{c}$                       &15.17& 9.69 &  3.15 \\ 
%     $\mathbf{a}t^{2}+\mathbf{b}t+\mathbf{c}$   &22.43& 11.05&  4.59\\ 
%     $e^{\mathbf{a}t}+\mathbf{c}$                   &28.15& 12.15&  5.33\\  \hline
%   \end{tabular}
% \end{table}

\vspace{-5pt}
\section{Conclusion}
We present a novel diffusion method named ADM to improve the generative abilities of DPMs. Different from prior approaches that only have an image-to-noise process, we incorporate the analytical image attenuation process into diffusion models, obtaining simultaneous image-to-zero and zero-to-noise processes. 
%which reduces the learning difficulty caused by the complex SDE. 
For the image to zero, we use an analytic transition function to model 
%Furthermore, we introduce the explicit transition probability to model 
the gradient of the image component, resulting in a diffusion process that enables the model to sample with an arbitrary step size. For unconditioned and conditioned image generation tasks, we show that ADM has consistently competitive performance under both long-step and few-step setups. 

%Our experiments demonstrate that DDM outperforms previous DPMs in terms of image synthesis and sampling compute budgets. 

\textbf{Future work and broader impact.} In principle, ADM can be combined with high-order ODE solvers to further improve the performance of few-step generation.  Yet we have not derived the formula of high-order solvers. %On the other hand, our experiments suggest that there is room for exploration in finding an optimal explicit function to fit the gradient of the image component, even though arbitrary explicit functions can be used. 
Like other deep generative models, ADM has the potential to transform many different fields and industries by generating new insights and positive social impact. % which makes the long-term social impacts dependent on its downstream applications.

% \input{appendix/exp_details}
% \nocite{*}

\bibliographystyle{IEEEtran} %.bst模板
\bibliography{ref.bib} %.bib文件名字
% \appendices
% \input{appendix/proof}
% \bibitem{ref1}
% {\it{Mathematics Into Type}}. American Mathematical Society. [Online]. Available: https://www.ams.org/arc/styleguide/mit-2.pdf

% \bibitem{ref2}
% T. W. Chaundy, P. R. Barrett and C. Batey, {\it{The Printing of Mathematics}}. London, U.K., Oxford Univ. Press, 1954.

% \bibitem{ref3}
% F. Mittelbach and M. Goossens, {\it{The \LaTeX Companion}}, 2nd ed. Boston, MA, USA: Pearson, 2004.

% \bibitem{ref4}
% G. Gr\"atzer, {\it{More Math Into LaTeX}}, New York, NY, USA: Springer, 2007.

% \bibitem{ref5}M. Letourneau and J. W. Sharp, {\it{AMS-StyleGuide-online.pdf,}} American Mathematical Society, Providence, RI, USA, [Online]. Available: http://www.ams.org/arc/styleguide/index.html

% \bibitem{ref6}
% H. Sira-Ramirez, ``On the sliding mode control of nonlinear systems,'' \textit{Syst. Control Lett.}, vol. 19, pp. 303--312, 1992.

% \bibitem{ref7}
% A. Levant, ``Exact differentiation of signals with unbounded higher derivatives,''  in \textit{Proc. 45th IEEE Conf. Decis.
% Control}, San Diego, CA, USA, 2006, pp. 5585--5590. DOI: 10.1109/CDC.2006.377165.

% \bibitem{ref8}
% M. Fliess, C. Join, and H. Sira-Ramirez, ``Non-linear estimation is easy,'' \textit{Int. J. Model., Ident. Control}, vol. 4, no. 1, pp. 12--27, 2008.

% \bibitem{ref9}
% R. Ortega, A. Astolfi, G. Bastin, and H. Rodriguez, ``Stabilization of food-chain systems using a port-controlled Hamiltonian description,'' in \textit{Proc. Amer. Control Conf.}, Chicago, IL, USA,
% 2000, pp. 2245--2249.

% \end{thebibliography}

\newpage

\vspace{11pt}

\begin{IEEEbiography}[{\includegraphics[width=1in,height=1.25in,clip,keepaspectratio]{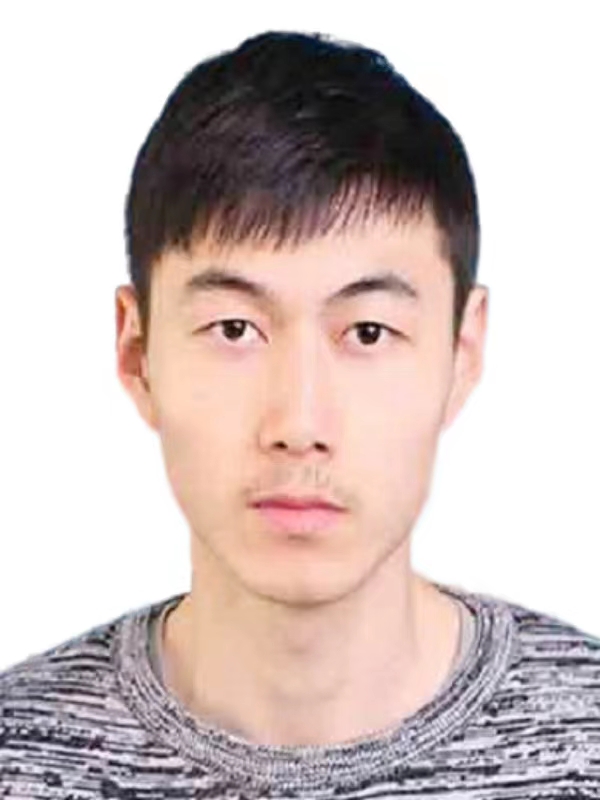}}]{Yuhang Huang}
received the bachelor’s degree from the University of Shanghai for Science and Technology, Shanghai, China, in 2019, and the master’s degree from Shanghai University, Shanghai, China, in 2022. He is currently pursuing the Ph.D. degree at the National University of Defense Technology, Changsha, China.

His research interests include computer vision, graphics, and generative models.
\end{IEEEbiography}

\vspace{11pt}

\begin{IEEEbiography}[{\includegraphics[width=1in,height=1.25in,clip,keepaspectratio]{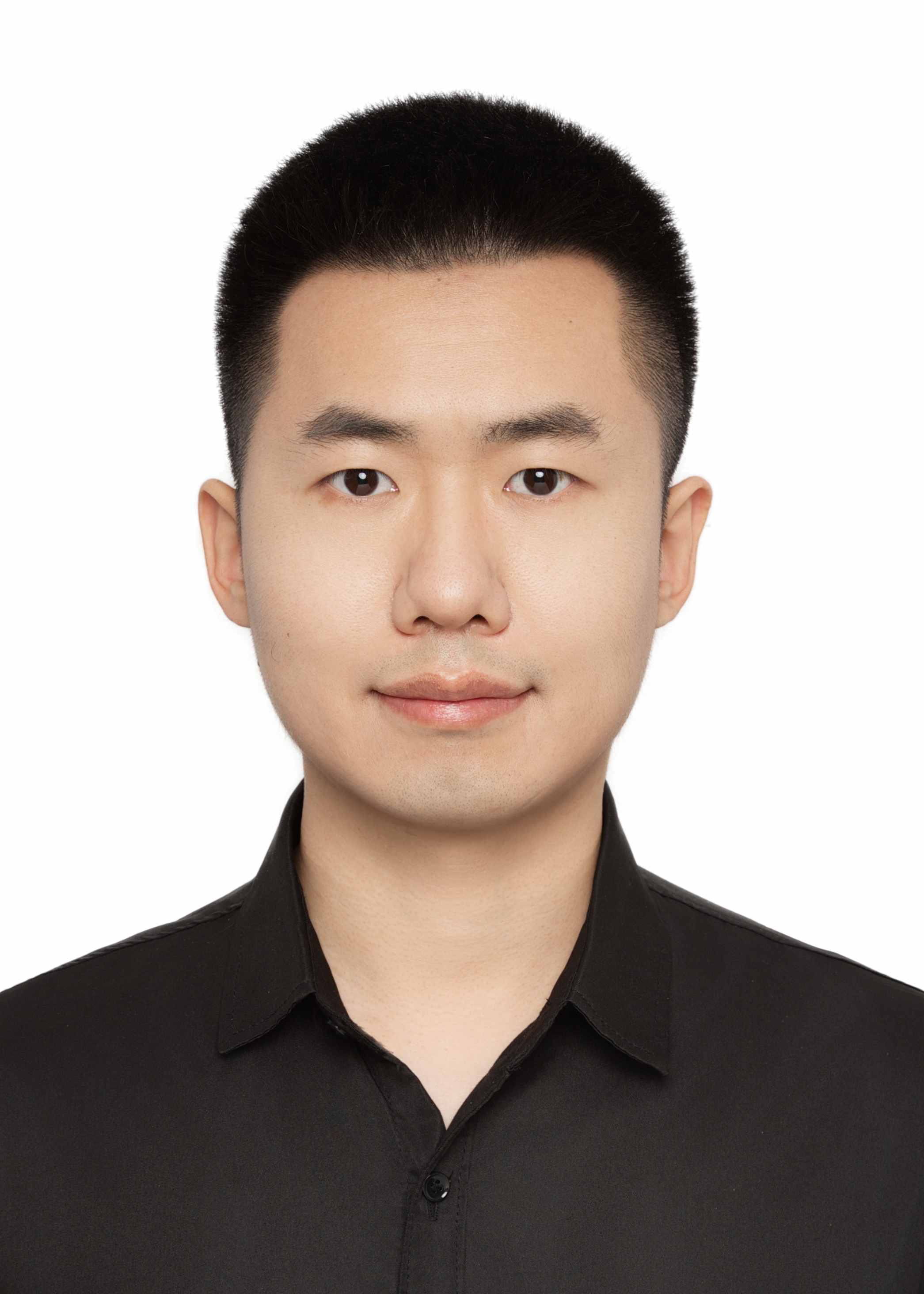}}]{Zheng Qin} 
received the PhD degree in computer science and technology from National University of Defense Technology (NUDT), China. He is currently an assistant professor with the Defense Innovation Institute, Academy of Military Sciences, China. His current research interests include 3D computer vision, 3D feature learning, shape modeling, and scene understanding.
% He has authored 10+ papers in journals and conferences such as IEEE T-PAMI, CVPR, ICCV, IJCAI, etc.
\end{IEEEbiography} 

\vspace{11pt}
\begin{IEEEbiography}[{\includegraphics[width=1in,height=1.25in,clip,keepaspectratio]{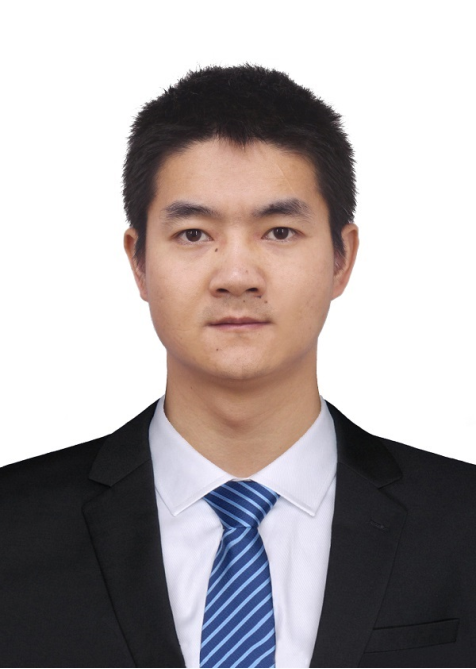}}]{Xinwang Liu}
received his PhD degree from National University of Defense Technology (NUDT), China. He is now Professor of School of Computer, NUDT. His current research interests include kernel learning and unsupervised feature learning. Dr. Liu has published 60+ peer-reviewed papers, including
those in highly regarded journals and conferences such as IEEE T-PAMI, IEEE T-KDE, IEEE T-IP, IEEE T-NNLS, IEEE T-MM, IEEE T-IFS, ICML, NeurIPS, ICCV, CVPR, AAAI, IJCAI, etc. He serves as the associated editor of Information Fusion Journal. More information can be found at \href{https://xinwangliu.github.io/}{https://xinwangliu.github.io/}.
\end{IEEEbiography}

\vspace{11pt}
\begin{IEEEbiography}[{\includegraphics[width=1in,height=1.25in,clip,keepaspectratio]{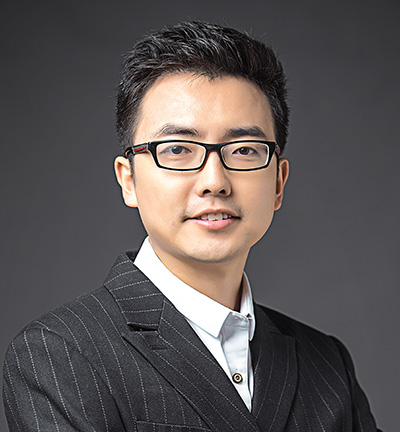}}]{Kai Xu} (Senior Member, IEEE) received the Ph.D. degree in computer science from the National University of Defense Technology (NUDT), Changsha,
China, in 2011. From 2008 to 2010, he worked as a Visiting Ph.D. degree with the GrUVi Laboratory, Simon Fraser University, Burnaby, BC, Canada. He is currently a Professor with the School of Computer Science, NUDT. He is also an Adjunct Professor with Simon Fraser University. His current research interests include data-driven shape analysis and modeling, and 3-D vision and robot perception and navigation.
\end{IEEEbiography}

\vfill

\end{document}

\title{Supplementary Material of Simultaneous Image to Zero and Zero to Noise: Diffusion Models with Analytical Image Attenuation}

\author{Yuhang Huang, Zheng Qin, Xinwang Liu,~\IEEEmembership{Senior Member,~IEEE}, Kai Xu$^{*}$\thanks{$*$ Corresponding author.},~\IEEEmembership{Senior Member,~IEEE}
% \thanks{Y. Huang, X. Liu and K. Xu are with School of Computer, National University of Defense Technology, Changsha, 410073, China (E-mail: {huangai, xinwangliu}\@nudt.edu.cn, kevin.kai.xu@gmail.com). Z. Qin is with Defense Innovation Institute, Academy of Military Sciences, Beijing, 100071, China.
}

\markboth{Journal of \LaTeX\ Class Files,~Vol.~14, No.~8, August~2021}%
{Shell \MakeLowercase{\textit{et al.}}: A Sample Article Using IEEEtran.cls for IEEE Journals}


% \maketitle
% \begin{figure*}[h]
%     \centering
%     % \includegraphics[width=1.\linewidth]{figures/teaser-cvpr.pdf}%\vspace{-10pt}
%     \includegraphics[width=1.\linewidth]{figures/figure1.pdf}%\vspace{-10pt}
%     \caption{High-quality images generated by the proposed DDMs under few-step settings. (a) 10-step unconditioned generation on the CIFAR-10 and CelebA-HQ-256 datasets. (b) 5-step conditioned tasks  %Generated samples on both unconditioned and conditioned tasks. \textbf{(a)}: DDMs can generate high-quality images on CIFAR-10 and CelebA-HQ-256 with only 10 steps. \textbf{(b)}: For conditioned tasks, DDMs only use 5 steps for generating high-quality results 
%     including saliency detection, image inpainting and super-resolution.}%\vspace{-15pt}
%     \label{fig:top}
% \end{figure*}

\maketitle

% \input{appendix/exp_details}
% \nocite{*}
\section{Proofs and Derivations}\label{supsec:1}
\subsection{Equivalence between ADM and previous diffusion process \cite{(2)ddpm}}\label{supsec:1.1}
The previous study \cite{(9)sde} has proved that the mapping from the image to noise can be formulated by a Stochastic Differential Equation (SDE):
\begin{equation}
     \mathrm{d}\mathbf{x}_{t} = f_{t}\mathbf{x}_{t}\mathrm{d}t + g_{t}\mathrm{d}\mathbf{w}_{t},\quad\mathbf{x}_{0} \sim q(\mathbf{x}_{0}),
     \label{sup_eq:1}
\end{equation}
where $f_{t}\mathbf{x}_{t}$ and $g_{t}$ represent the drift term and diffusion term of the It\^{o} diffusion process \cite{(43)ito1996diffusion}.
Our forward diffusion process is directly described by the It\^{o} integral that can be formulated by:
\begin{equation}
     \mathbf{x}_{t} = \mathbf{x}_{0} + \int_{0}^{t} {\mathbf{h}_{t}\mathrm{d}t} + \int_{0}^{t} {\mathrm{d}\mathbf{w}_{t}},
     \quad\mathbf{x}_{0} \sim q(\mathbf{x}_{0}).
     \label{sup_eq:2}
\end{equation}
Taking the derivative of the above formula, we get a similar differential form to Eq.~\ref{sup_eq:1}:
\begin{equation}
     \mathrm{d}\mathbf{x}_{t} = \mathbf{h}_{t}\mathrm{d}t + \mathrm{d}\mathbf{w}_{t},\quad\mathbf{h}_{t}\sim \int_{0}^{t}\mathbf{h}_{t}\mathrm{d}t+\mathbf{x}_{0}=\mathbf{0},\quad\mathbf{x}_{0}\sim q(\mathbf{x}_{0}),
     \label{sup_eq:3}
\end{equation}
Comparing Eq.~\ref{sup_eq:1} and Eq.~\ref{sup_eq:3}, we can prove that the two formulas are equal when $f_{t}\mathbf{x}_{t}=h_{t}$ and $g_{t}=1$. Therefore, the proposed diffusion process is equivalent to previous diffusion processes described by Eq.~\ref{sup_eq:1}. Actually, our diffusion formula is a special form of Eq.~\ref{sup_eq:1} that they both can describe the mapping from the image to noise. Differently, the proposed diffusion process allows the model to learn the image and noise components independently. 
We incorporate the analytical image attenuation process into diffusion models and propose the simultaneous image to zero and zero to noise, reducing the learning difficulty and enabling an approximate analytical solution for the reversed process.
% We decouple the complex mapping from the original image to noise into two relatively simpler processes, reducing the learning difficulty and enabling using an analytical transition probability to model the image component. 

\subsection{Proof of Training Objective}\label{supsec:1.2}
We can derive the training objective of ADM by maximizing the evidence lower bound of the log-likelihood $\log{p(\mathbf{x}_{0})}$. Considering the continuous-time Markov chain, $\log{p(\mathbf{x}_{0})}$ is represented by:
% \resizebox{\linewidth}{!}
{
\begin{small}
\begin{equation}
    \begin{aligned}
     \log{p(\mathbf{x}_{0})} &= \log{\int_{}^{} {p(\mathbf{x}_{0:1})\mathrm{d} \mathbf{x}_{\Delta t:1}} }, \Delta t\rightarrow 0^{+} \\
        &= \log{\int_{}^{} \frac{p(\mathbf{x}_{0:1})q(\mathbf{x}_{\Delta t:1}|\mathbf{x}_{0})}{q(\mathbf{x}_{\Delta t:1}|\mathbf{x}_{0})}\mathrm{d} \mathbf{x}_{\Delta t:1}} \\
        &= \log{\mathbb{E}_{q}[\frac{p(\mathbf{x}_{0:1})}{q(\mathbf{x}_{\Delta t:1}|\mathbf{x}_{0})}]} \\
        &\geq \mathbb{E}_{q}[\log{\frac{p(\mathbf{x}_{0:1})}{q(\mathbf{x}_{\Delta t:1}|\mathbf{x}_{0})}}] \quad\# \text{Jensen’s inequality}\\
        &= \mathbb{E}_{q}[\log{\frac{ p(\mathbf{x}_{1})\prod_{t=\Delta t}^{1}{p_{\boldsymbol{\theta}}(\mathbf{x}_{t-\Delta t}|\mathbf{x}_{t})} }  
           {\prod_{t=\Delta t}^{1}{q(\mathbf{x}_{t}|\mathbf{x}_{t-\Delta t})}} }] \\
        &= \mathbb{E}_{q}[\log{\frac{ p(\mathbf{x}_{1})\prod_{t=\Delta t}^{1}{p_{\boldsymbol{\theta}}(\mathbf{x}_{t-\Delta t}|\mathbf{x}_{t})} }  
           {\prod_{t=\Delta t}^{1}{q(\mathbf{x}_{t}|\mathbf{x}_{t-\Delta t}, \mathbf{x}_{0})}} }]\\
        &= \mathbb{E}_{q}[\log{\frac{ p(\mathbf{x}_{1})\prod_{t=\Delta t}^{1}{p_{\boldsymbol{\theta}}(\mathbf{x}_{t-\Delta t}|\mathbf{x}_{t})}q(\mathbf{x}_{t-\Delta t}|\mathbf{x}_{0}) }  
           {\prod_{t=\Delta t}^{1}{q(\mathbf{x}_{t-\Delta t}|\mathbf{x}_{t}, \mathbf{x}_{0})} q(\mathbf{x}_{t}|\mathbf{x}_{0}) } }]\\
        &= \mathbb{E}_{q}[\log{\frac{ p(\mathbf{x}_{1})\prod_{t=\Delta t}^{1}{p_{\boldsymbol{\theta}}(\mathbf{x}_{t-\Delta t}|\mathbf{x}_{t})} }  
           {\prod_{t=\Delta t}^{1}{q(\mathbf{x}_{t-\Delta t}|\mathbf{x}_{t}, \mathbf{x}_{0})} }}\\
        &+ \log{\frac{ \prod_{t=\Delta t}^{1}{q(\mathbf{x}_{t-\Delta t}|\mathbf{x}_{0})} }{\prod_{t=\Delta t}^{1}{q(\mathbf{x}_{t}|\mathbf{x}_{0})} }}] \\
        &= \mathbb{E}_{q}[\log{\frac{ p(\mathbf{x}_{1})\prod_{t=\Delta t}^{1}{p_{\boldsymbol{\theta}}(\mathbf{x}_{t-\Delta t}|\mathbf{x}_{t})} }  
           {\prod_{t=\Delta t}^{1}{q(\mathbf{x}_{t-\Delta t}|\mathbf{x}_{t}, \mathbf{x}_{0})} }}
           + \log{\frac{1}{q(\mathbf{x}_{1}|\mathbf{x}_{0})}}] \\
        &= \mathbb{E}_{q}[\log{\frac{p(\mathbf{x}_{1})}{q(\mathbf{x}_{1}|\mathbf{x}_{0})}} + \log{\frac{\prod_{t=\Delta t}^{1}{p_{\boldsymbol{\theta}}(\mathbf{x}_{t-\Delta t}|\mathbf{x}_{t})} }  
           {\prod_{t=\Delta t}^{1}{q(\mathbf{x}_{t-\Delta t}|\mathbf{x}_{t}, \mathbf{x}_{0})} }}] \\
        &= \mathbb{E}_{q}[ \log{\frac{p(\mathbf{x}_{1})}{q(\mathbf{x}_{1}|\mathbf{x}_{0})}} ] \\
        &- \sum_{t=\Delta t}^{1}\mathbb{E}_{q}[D_{KL}(q(\mathbf{x}_{t-\Delta t}|\mathbf{x}_{t}, \mathbf{x}_{0})||p_{\boldsymbol{\theta}}(\mathbf{x}_{t-\Delta t}|\mathbf{x}_{t}))]
    \end{aligned}
    \label{sup_eq:4}
\end{equation}
\end{small}
}
The first term in the equation can be interpreted as the prior matching term, which ensures that the original data can be transformed into the noise distribution when $t=1$. It has been demonstrated in Section 3 of the main paper that our forward process satisfies this requirement, eliminating the need to construct a loss function for this term. 
On the other hand, the latter term represents the denoising matching term and its purpose is to ensure consistency in the distribution at $\mathbf{x}_{t}$ from both the forward and reversed processes. To achieve this, we aim to learn the desired denoising transition step $p_{\boldsymbol{\theta}}(\mathbf{x}_{t-\Delta t}|\mathbf{x}_{t})$ as an approximation to the tractable ground-truth denoising transition step $q(\mathbf{x}_{t-\Delta t}|\mathbf{x}_{t}, \mathbf{x}_{0})$. Minimizing this term entails achieving the closest possible match between the two denoising steps, as quantified by their KL Divergence.
In fact, the ground-truth denoising transition step $q(\mathbf{x}_{t-\Delta t}|\mathbf{x}_{t}, \mathbf{x}_{0})$ is still a normal distribution. Therefore, minimizing the KL Divergence is equivalent to minimizing the error between their mean and variance, which means that we can use the MSE (mean squared error) function to formulate the training object.

We first prove the normality of $q(\mathbf{x}_{t-\Delta t}|\mathbf{x}_{t}, \mathbf{x}_{0})$. Giving the forward transition probability $q(\mathbf{x}_{t}|\mathbf{x}_{0}) = \mathcal{N}(\mathbf{x}_{t}; \mathbf{x}_{0}+\mathbf{H}_{t}, t\mathbf{I})$, we have: %and it, so we do not but transfer . We first prove $$ . 
\begin{footnotesize}
\begin{equation}
    \begin{aligned}
        q(\mathbf{x}_{t-\Delta t}|\mathbf{x}_{t}, \mathbf{x}_{0}) &= \frac{q(\mathbf{x}_{t}|\mathbf{x}_{t-\Delta t}, \mathbf{x}_{0})q(\mathbf{x}_{t-\Delta t}|\mathbf{x}_{0})} {q(\mathbf{x}_{t}|\mathbf{x}_{0})} \\
        &=\frac{q(\mathbf{x}_{t}|\mathbf{x}_{t-\Delta t})q(\mathbf{x}_{t-\Delta t}|\mathbf{x}_{0})} {q(\mathbf{x}_{t}|\mathbf{x}_{0})} \\
        &= \frac{q(\mathbf{x}_{t}|\mathbf{x}_{t-\Delta t}) \mathcal{N}(\mathbf{x}_{t-\Delta t}; \mathbf{x}_{0}+\mathbf{H}_{t-\Delta t}, (t-\Delta t)\mathbf{I})}{\mathcal{N}(\mathbf{x}_{t}; \mathbf{x}_{0}+\mathbf{H}_{t}, t\mathbf{I})}.
    \end{aligned}
     \label{sup_eq:5}
\end{equation}
\end{footnotesize}
Here, we can easily prove that $q(\mathbf{x}_{t}|\mathbf{x}_{t-\Delta t})$ is also a normal distribution by the following derivation:
\begin{equation}
    \begin{aligned}
        \mathbf{x}_{t} &= \mathbf{x}_{0} + \mathbf{H}_{t} + \sqrt{t}\boldsymbol{\epsilon}, \boldsymbol{\epsilon}\sim \mathcal{N}(\mathbf{0}, \mathbf{I}) \\
        &= \mathbf{x}_{0} + \mathbf{H}_{t} - \mathbf{H}_{t-\Delta t} + \mathbf{H}_{t-\Delta t} + \sqrt{t-\Delta t}\boldsymbol{\epsilon}_{1} + \sqrt{\Delta t}\boldsymbol{\epsilon}_{2}\\
        &= \mathbf{x}_{0} + \mathbf{H}_{t-\Delta t} + \sqrt{t-\Delta t}\boldsymbol{\epsilon}_{1} + \mathbf{H}_{t} - \mathbf{H}_{t-\Delta t} + \sqrt{\Delta t}\boldsymbol{\epsilon}_{2} \\
        &=\mathbf{x}_{t-\Delta t} + \mathbf{H}_{t} - \mathbf{H}_{t-\Delta t} + \sqrt{\Delta t}\boldsymbol{\epsilon}_{2},
    \end{aligned}
     \label{sup_eq:6}
\end{equation}
where, $\boldsymbol{\epsilon}_{1}, \boldsymbol{\epsilon}_{2}\sim \mathcal{N}(\mathbf{0}, \mathbf{I})$.
Thus, $q(\mathbf{x}_{t}|\mathbf{x}_{t-\Delta t})$ is the normal distribution with mean $\mathbf{x}_{t-\Delta t} + \mathbf{H}_{t} - \mathbf{H}_{t-\Delta t}$ and variance $\Delta t\mathbf{I}$:
\begin{equation}
    q(\mathbf{x}_{t}|\mathbf{x}_{t-\Delta t})=\mathcal{N}(\mathbf{x}_{t}; \mathbf{x}_{t-\Delta t} + \mathbf{H}_{t} - \mathbf{H}_{t-\Delta t}, \Delta t\mathbf{I}).
     \label{sup_eq:7}
\end{equation} 
Substituting Eq.~\ref{sup_eq:7} into Eq.~\ref{sup_eq:5}, we have:
\begin{footnotesize}
\begin{equation}
    \begin{aligned}
        q(\mathbf{x}_{t-\Delta t}|\mathbf{x}_{t}, \mathbf{x}_{0}) &= \frac{q(\mathbf{x}_{t}|\mathbf{x}_{t-\Delta t}) \mathcal{N}(\mathbf{x}_{t-\Delta t}; \mathbf{x}_{0}+\mathbf{H}_{t-\Delta t}, (t-\Delta t)\mathbf{I})}{\mathcal{N}(\mathbf{x}_{t}; \mathbf{x}_{0}+\mathbf{H}_{t}, t\mathbf{I})}\\
        &\propto \exp\{-\frac{1}{2}[\frac{(\mathbf{x}_{t}-\mathbf{x}_{t-\Delta t}-\mathbf{H}_{t}+\mathbf{H}_{t-\Delta t})^{2}}{\Delta t} \\&+\frac{(\mathbf{x}_{t-\Delta t}-\mathbf{x}_{0}-\mathbf{H}_{t-\Delta t})^{2}}{t-\Delta t}]\} \\
        &= \exp\{-\frac{1}{2}[\frac{t}{\Delta t(t-\Delta t)}\mathbf{x}_{t-\Delta t}^{2} \\
        &+ \frac{\Delta t\mathbf{x}_{0}+(t-\Delta t)(\mathbf{x}_{t}-\mathbf{H}_{t})+t\mathbf{H}_{t-\Delta t}} {\Delta t(t-\Delta t)}\\&\cdot(-2\mathbf{x}_{t-\Delta t}) + C]\},
    \end{aligned}
     \label{sup_eq:8}
\end{equation}
\end{footnotesize}
where $C$ is the term not related to $\mathbf{x}_{t-\Delta t}$. Eq.~\ref{sup_eq:8} contains $\mathbf{x}_{0}$ which is known in the forward process, and we can directly get the prior $\mathbf{x}_{0}$ from the forward transition probability $q(\mathbf{x}_{t}|\mathbf{x}_{0})$:
\begin{equation}
    \mathbf{x}_{0}=\mathbf{x}_{t} - \mathbf{H}_{t} - \sqrt{t}\boldsymbol{\epsilon}.
     \label{sup_eq:9}
\end{equation}
Comparing Eq.~\ref{sup_eq:8} with the standard normal distribution and substituting Eq.~\ref{sup_eq:9} into Eq.~\ref{sup_eq:8}, we can obtain the mean $\mathbf{\widetilde{u}}$ and variance $\widetilde{\sigma}^{2}$ of $q(\mathbf{x}_{t-\Delta t}|\mathbf{x}_{t}, \mathbf{x}_{0})$:
\begin{equation}
    \begin{split}
     &\mathbf{\widetilde{u}} = \mathbf{x}_{t}+\mathbf{H}_{t-\Delta t}-\mathbf{H}_{t}-\frac{\Delta t}{\sqrt{t}}\boldsymbol{\epsilon},\\
     &\widetilde{\sigma}^{2}=\frac{\Delta t(t-\Delta t)}{t}.
    \end{split}
    \label{sup_eq:10}
\end{equation}
Thus, the proof of $q(\mathbf{x}_{t-\Delta t}|\mathbf{x}_{t}, \mathbf{x}_{0})$ is complete.

We use a neural network with parameters $\boldsymbol{\theta}$ to parameterize $p_{\boldsymbol{\theta}}(\mathbf{x}_{t-\Delta t}|\mathbf{x}_{t})$. In practice, the variance $\widetilde{\sigma}^{2}$ only depends on $t$ and $\Delta t$ so we only need to parameterize the mean. From Eq.~\ref{sup_eq:4}, we can optimize $\boldsymbol{\theta}$ by minimizing the KL Divergence:
\begin{equation}
    \begin{aligned}
        &\min\limits_{\boldsymbol{\theta}} D_{KL}(q(\mathbf{x}_{t-\Delta t}|\mathbf{x}_{t}, \mathbf{x}_{0})||p_{\boldsymbol{\theta}}(\mathbf{x}_{t-\Delta t}|\mathbf{x}_{t})) \\
        =&\min\limits_{\boldsymbol{\theta}} D_{KL}(\mathcal{N}(\mathbf{x}_{t-\Delta t}; \mathbf{\widetilde{u}}, \widetilde{\sigma}^{2}\mathbf{I})||\mathcal{N}(\mathbf{x}_{t-\Delta t}; \mathbf{u}_{\boldsymbol{\theta}}, \widetilde{\sigma}^{2}\mathbf{I})) \\
        =&\min\limits_{\boldsymbol{\theta}} \frac{1}{2} (\mathbf{u}_{\boldsymbol{\theta}}- \mathbf{\widetilde{u}})^{T} (\widetilde{\sigma}^{2}\mathbf{I})^{-1} (\mathbf{u}_{\boldsymbol{\theta}}- \mathbf{\widetilde{u}}) \\
        =&\min\limits_{\boldsymbol{\theta}} \frac{1}{2\widetilde{\sigma}^{2}}\Vert\mathbf{u}_{\boldsymbol{\theta}}- \mathbf{\widetilde{u}}\Vert^{2}.
    \end{aligned}
     \label{sup_eq:11}
\end{equation}
As $\mathbf{u}_{\boldsymbol{\theta}}$ is also conditioned on $\mathbf{x}_{t}$, we can match $\mathbf{\widetilde{u}}$ closely by setting it to the following form:
\begin{equation}
    \mathbf{u}_{\boldsymbol{\theta}}=\mathbf{x}_{t}+\mathbf{H_{\boldsymbol{\theta}}}_{t-\Delta t}-\mathbf{H_{\boldsymbol{\theta}}}_{t}-\frac{\Delta t}{\sqrt{t}}\boldsymbol{\epsilon}_{\boldsymbol{\theta}}.
     \label{sup_eq:12}
\end{equation}
%where $$ and are parameterized by the neural network that seeks to predict x0 from noisy image $\mathbf{x}_{t}$ and time $t$. 
Then, the optimization problem simplifies to:
\begin{equation}
    \begin{aligned}
        &\min\limits_{\boldsymbol{\theta}} D_{KL}(q(\mathbf{x}_{t-\Delta t}|\mathbf{x}_{t}, \mathbf{x}_{0})||p_{\boldsymbol{\theta}}(\mathbf{x}_{t-\Delta t}|\mathbf{x}_{t})) \\
        =&\min\limits_{\boldsymbol{\theta}} \frac{1}{2\widetilde{\sigma}^{2}}\Vert(\mathbf{x}_{t}+\mathbf{H_{\boldsymbol{\theta}}}_{t-\Delta         t}-\mathbf{H_{\boldsymbol{\theta}}}_{t}-\frac{\Delta t}{\sqrt{t}}\boldsymbol{\epsilon}_{\boldsymbol{\theta}}) \\&- 
            (\mathbf{x}_{t}+\mathbf{H}_{t-\Delta t}-\mathbf{H}_{t}-\frac{\Delta t}{\sqrt{t}}\boldsymbol{\epsilon})\Vert^{2}\\
        =&\min\limits_{\boldsymbol{\theta}} \frac{1}{2\widetilde{\sigma}^{2}}\Vert\int_{t}^{t-\Delta t}{\mathbf{h_{\boldsymbol{\theta}}}_{t} \mathrm{d}t} - \int_{t}^{t-\Delta t} {\mathbf{h}_{t}\mathrm{d}t} + \frac{\Delta t}{\sqrt{t}}(\boldsymbol{\epsilon}-\boldsymbol{\epsilon}_{\boldsymbol{\theta}})\Vert^{2} \\
        =&\min\limits_{\boldsymbol{\theta}} \frac{1}{2\widetilde{\sigma}^{2}}\Vert\int_{t}^{t-\Delta t}{\mathbf{h_{\boldsymbol{\theta}}}_{t}-\mathbf{h}_{t}\mathrm{d}t}  + \frac{\Delta t}{\sqrt{t}}(\boldsymbol{\epsilon}-\boldsymbol{\epsilon}_{\boldsymbol{\theta}})\Vert^{2}\\
        =&:\min\limits_{\boldsymbol{\theta}} \frac{1}{2\widetilde{\sigma}^{2}}\Vert{\mathbf{h_{\boldsymbol{\theta}}}_{t}-\mathbf{h}_{t}}\Vert^{2}  + \frac{1}{2\widetilde{\sigma}^{2}}\Vert\frac{\Delta t}{\sqrt{t}}(\boldsymbol{\epsilon}_{\boldsymbol{\theta}}-\boldsymbol{\epsilon})\Vert^{2}\\
        =&\min\limits_{\boldsymbol{\theta}} \frac{1}{2\widetilde{\sigma}^{2}}\Vert{\mathbf{h_{\boldsymbol{\theta}}}_{t}-\mathbf{h}_{t}}\Vert^{2}  + \frac{\Delta t^{2}}{2\widetilde{\sigma}^{2}t}\Vert\boldsymbol{\epsilon}_{\boldsymbol{\theta}}-\boldsymbol{\epsilon}\Vert^{2}.
    \end{aligned}
     \label{sup_eq:13}
\end{equation}
As $\mathbf{h}_{t}$ is determined by its hyper-parameters $\boldsymbol{\phi}$ that can be solved using $\mathbf{x}_{0} + \int_{0}^{1} {\mathbf{h}_{t}\mathrm{d}t}=\mathbf{0}$, we can directly parameterize $\boldsymbol{\phi}$ using $\boldsymbol{\phi}_{\boldsymbol{\theta}}$. Thus, the training objective is formulated by:
\begin{equation}
    \begin{aligned}
        &\min\limits_{\boldsymbol{\theta}} D_{KL}(q(\mathbf{x}_{t-\Delta t}|\mathbf{x}_{t}, \mathbf{x}_{0})||p_{\boldsymbol{\theta}}(\mathbf{x}_{t-\Delta t}|\mathbf{x}_{t})) \\
        =&\min\limits_{\boldsymbol{\theta}} \frac{1}{2\widetilde{\sigma}^{2}}\Vert{\boldsymbol{\phi}_{\boldsymbol{\theta}}-\boldsymbol{\phi}}\Vert^{2}  + \frac{\Delta t^{2}}{2\widetilde{\sigma}^{2}t}\Vert\boldsymbol{\epsilon}_{\boldsymbol{\theta}}-\boldsymbol{\epsilon}\Vert^{2}.
    \end{aligned}
     \label{sup_eq:14}
\end{equation}
The proof of our training objective is completed. In essential, $\boldsymbol{\phi}$ represents the original image component while $\boldsymbol{\epsilon}$ can be seen as the noise distribution. Therefore, the training objective allows the model to learn the image and the noise components independently.

\subsection{Derivation of Few-step Sampling Formula}\label{supsec:1.3}
Different from previous diffusion probabilistic models (DPMs), the proposed ADM naturally enables few-step sampling. Following Eq.\ref{sup_eq:6}, we have:
\begin{equation}
    \begin{aligned}
        \mathbf{x}_{t} &= \mathbf{x}_{0} + \mathbf{H}_{t} + \sqrt{t}\boldsymbol{\epsilon}, \boldsymbol{\epsilon}\sim \mathcal{N}(\mathbf{0}, \mathbf{I}) \\
        &= \mathbf{x}_{0} + \mathbf{H}_{t} - \mathbf{H}_{t-s} + \mathbf{H}_{t-s} + \sqrt{t-s}\boldsymbol{\epsilon}_{1} \\
        &+ \sqrt{s}\boldsymbol{\epsilon}_{2}, \boldsymbol{\epsilon}_{1}, \boldsymbol{\epsilon}_{2}\sim \mathcal{N}(\mathbf{0}, \mathbf{I})\\
        &= \mathbf{x}_{0} + \mathbf{H}_{t-s} + \sqrt{t-s}\boldsymbol{\epsilon}_{1} + \mathbf{H}_{t} - \mathbf{H}_{t-s} + \sqrt{s}\boldsymbol{\epsilon}_{2} \\
        &=\mathbf{x}_{t-s} + \mathbf{H}_{t} - \mathbf{H}_{t-s} + \sqrt{s}\boldsymbol{\epsilon}_{2}. 
    \end{aligned}
     \label{sup_eq:15}
\end{equation}
Thus, we can give the transition probability $q(\mathbf{x}_{t}|\mathbf{x}_{t-s})$ for an arbitrary step size $s$:
\begin{equation}
    q(\mathbf{x}_{t}|\mathbf{x}_{t-s})=\mathcal{N}(\mathbf{x}_{t}; \mathbf{x}_{t-s} + \mathbf{H}_{t} - \mathbf{H}_{t-s}, s\mathbf{I}).
     \label{sup_eq:16}
\end{equation}
With Eq.~\ref{sup_eq:16}, we can easily derive the mean and variance of $q(\mathbf{x}_{t-s}|\mathbf{x}_{t}, \mathbf{x}_{0})$ following Eq.~\ref{sup_eq:8}:
\begin{equation}
    \begin{split}
     &q(\mathbf{x}_{t-s}|\mathbf{x}_{t}, \mathbf{x}_{0}) \propto \exp\{-\frac{(\mathbf{x}_{t-s}-\mathbf{\widetilde{u}})^{2}}{2\widetilde{\sigma}^{2}\mathbf{I}}\}, \\
     &\mathbf{\widetilde{u}} = \mathbf{x}_{t}+\mathbf{H}_{t-s}-\mathbf{H}_{t}-\frac{s}{\sqrt{t}}\boldsymbol{\epsilon},\\
     &\widetilde{\sigma}^{2}=\frac{s(t-s)}{t}.
     \end{split}
     \label{sup_ep:17}
\end{equation}
Eq.~\ref{sup_ep:17} means ADM can sample with any step size even using the largest step size $s=1$ for one-step generations. However, it is challenging to estimate an exactly accurate $\boldsymbol{\phi}$ at the initial time, and one-step generation causes the variance $\widetilde{\sigma}^{2}$ to be zero. This results in the generated images appearing somewhat blurry and lacking diversity. Therefore, we still sample iteratively but use a much larger step size than previous DPMs, reducing the sampling steps from 1000 to 10 drastically. We commence at $t=1$ and take uniform steps of size $s$ until we reach $t=0$.

\bibliographystyle{IEEEtran} %.bst模板
\bibliography{ref.bib} %.bib文件名字

\vfill